\PassOptionsToPackage{numbers}{natbib}
\documentclass{article}

\usepackage[preprint]{neurips_2025}

\usepackage[utf8]{inputenc} 
\usepackage[T1]{fontenc}    
\usepackage{hyperref}       
\usepackage{url}            
\usepackage{booktabs}       
\usepackage{amsfonts}       
\usepackage{nicefrac}       
\usepackage{microtype}      
\usepackage{xcolor}         

\usepackage{mathtools}

\usepackage{float}
\usepackage{dblfloatfix}    
\usepackage{bm}
\usepackage{lipsum}
\usepackage{amsmath}
\usepackage{comment}
\usepackage{colortbl}

\usepackage{caption}
\usepackage{graphicx}
\usepackage{hhline}
\usepackage{tabularray}
\usepackage[ruled,vlined]{algorithm2e}
\usepackage{multirow}

\usepackage{minitoc}

\usepackage{subcaption}

\usepackage{amssymb}
\usepackage{pifont}
\newcommand{\cmark}{\ding{51}}%
\newcommand{\xmark}{\ding{55}}%

\usepackage[shortlabels]{enumitem}

\def\x{{\bf x}}
\def\z{{\bf z}}

\def\y{{\bf y}}
\def\Y{{\cal Y}}

\def \i {{_{i}}}
\def \t {{^{t}}}

\def \D {{\mathcal{D}}}
\def \mubold {{\boldsymbol{\mu}}}
\def \sigmabold {{\boldsymbol{\sigma}}}
\def \epsilonbold {{\boldsymbol{\epsilon}}}

\newcommand{\Tau}{\mathcal{T}}

\def\promptfn{{q}}
\def\incrmodel{{h}}
\def\incrmodelparams{{\psi}}

\DeclareMathOperator{\Tr}{Tr}

\newcommand{\mysumyy}{
	\sum_{\substack{\y,\y' \in \Y^{1:t} \\ \y \neq \y'}}
}

\newcommand{\mysumy}{
	\sum_{\substack{\y' \in \Y^{1:t} \\ \y' \neq \y}}
}

\author{
	Victor Enescu \quad\quad\quad\quad\quad Hichem Sahbi \\
	Sorbonne University, CNRS, LIP6, F-75005, Paris, France \\
	\texttt{victor.enescu@lip6.fr, hichem.sahbi@lip6.fr}
}

\begin{document}

\title{Frugal Incremental Generative Modeling using Variational Autoencoders}

\maketitle

\begin{abstract}
Continual or incremental learning holds tremendous potential in deep learning with different challenges including catastrophic forgetting. The advent of powerful foundation and generative models has propelled this paradigm even further, making it one of the most viable solution to train these models. However, one of the persisting issues lies in the increasing volume of data particularly with replay-based methods. This growth introduces challenges with scalability since continuously expanding data becomes increasingly demanding as the number of tasks grows. In this paper, we attenuate this issue by devising a novel replay-free incremental learning model based on Variational Autoencoders (VAEs). The main contribution of this work includes (i) a novel incremental generative modelling, built upon a well designed multi-modal latent space, and also (ii) an orthogonality criterion that mitigates catastrophic forgetting of the learned VAEs. The proposed method considers two variants of these VAEs: static and dynamic with no (or at most a controlled) growth in the number of parameters. Extensive experiments show that our method is (at least) an order of magnitude more ``memory-frugal'' compared to the closely related works while achieving SOTA accuracy scores.
\end{abstract}

\section{Introduction}
\label{sec:intro}
Deep neural networks (DNNs) have achieved outstanding results across various domains ranging from classification to generative modeling, often exceeding human performance, particularly when trained with substantial labeled data. However, unlike humans' inherent ability for continuous lifelong learning, DNNs struggle to learn new tasks without forgetting previous ones. This challenge, known as \emph{catastrophic forgetting}~\cite{cf}, hinders the deployment of large-scale incremental DNNs and stems directly from gradient-based optimization. Specifically, during the acquisition of a new task, weights optimized for previous data distributions are overwritten, leading to forgetting, a phenomenon exacerbated by sequential task learning. This paradigm, which consists in training through successive data streams or tasks, is known as Continual Learning (CL)~\cite{thrun1995lifelong, ring1997child, parisi2019continual}.\\
\indent The advent of transformer-based foundation models~\cite{transformer, transformer_vision} has ushered in a new era for continual learning, providing powerful backbones that narrow the performance gap with humans, particularly in zero/few-shot scenarios. This shift leverages rich representations, moving away from training limited feature extractors towards innovative adaptation of models like CLIP~\cite{clip, clip_scaling}. While standard fine-tuning~\cite{clip_finetune, clip_finetune_improve,clip_one_shot_forget} remains computationally intensive on large foundation models, alternatives inspired by language models, such as low-rank adaptation~\cite{lora} and adapter layers~\cite{adapter}, offer more efficient solutions. Notably, prompt learning~\cite{coop, cocoop, attriclip, cgil} for the CLIP text encoder, using additive parameters or networks aligned with image features, demonstrates huge potential. These parameter-efficient fine-tuning (PEFT) methods freeze the core foundation model, and adjust only a small fraction of dynamic parameters, either shared or class-specific. A fundamental challenge in these  models is task interference~\cite{star-prompt}; in other words, the absence of previous class data, when training new ones, may induce class overlap leading to severe performance degradation. Simple solutions, such as storing gaussian parameters for class-specific resampling~\cite{fetril, il2a, pass}, offer a minimal memory footprint. However, their limited expressiveness restricts the achievable accuracy. Consequently, alternative approaches employ generative models~\cite{dm_cl_features, gan_feature_replay, cgil} to synthesize data, bypassing explicit memory storage. Yet, many of these methods rely on class or task-dependent generative models~\cite{dm_cl_features,cgil, brain_inspired_replay_vae}, or achieve self-replay to mitigate performance decline~\cite{gan_feature_replay, nf_cl_2}. The resulting linear increase in both memory and computational demands renders these solutions impractical for realistic, long-term incremental learning scenarios. \\
\indent In order to address the foregoing challenges, we present a novel incremental learning model based on conditional Variational Autoencoders (VAEs)~\cite{vae}. Our model circumvents costly self-replay by instead leveraging null-space optimization via gradient projection~\cite{adamnscl}. Furthermore, the VAE is conditioned on multiple separated gaussians, incrementally learned in the latent space using a fixed-point iteration (FPI) framework with a Kullback-Leibler Divergence (KLD) repulsion criterion between gaussian pairs. This design ensures high accuracy on prior classes while learning new ones. The multi-gaussian conditioning in the latent space also reduces forgetting and label errors (common in generative replay), as gaussians are inherently non-overlapping. Note that VAEs are well-suited for generative incremental learning due to their low-dimensional latent spaces, enabling efficient gaussian parameter estimation in data-scarce scenarios. The efficacy of these contributions is validated through comprehensive experiments on four established incremental learning benchmarks, including fine-grained datasets like Cars196 and CUB200. These experiments demonstrate (i) superior performance over baselines and (ii) outperformance against memory replay methods (see Section \ref{sec:experiments}), highlighting the strong generalization ability of the proposed model. \\
Considering all the aforementioned issues, the main contributions of this paper include: 
\begin{enumerate}[label=(\roman*), topsep=-0.1pt]
	\item A new incremental learning framework that learns variational autoencoders. The proposed method learns new classes without performance degradation on previous ones. The proposed VAE is replay-free and relies on null space optimization in order to learn new tasks without increasing its memory footprint while maintaining a high accuracy. 
	
	\item \textcolor{black}{A novel conditioning mechanism for VAEs, that results into highly accurate labels when generating data, and is based on a fixed point iteration framework applied in the latent space.}
	
	\item  Extensive experiments through different benchmarks showing the outperformance of our proposed incremental learning method against state-of-the-art related work; for instance, on the Cars196 benchmark, our method overtakes previous approaches by up to 1.3 accuracy points.
\end{enumerate}

\section{Related Work}
\paragraph{Continual Learning.} Three different categories of continual learning methods exist in the literature: \emph{(1) dynamic architectures} \cite{dynamic_cl_1, dynamic_cl_2, dynamic_cl_3, dynamic_cl_4} which allocate different subnetwork parameters for different tasks, \emph{(2) regularization-based} methods \cite{reg_cl_1, reg_cl_2, reg_cl_3, reg_cl_4} that balance plasticity and rigidity in order to prevent significant updates of important parameters from previous tasks and thereby avoid catastrophic forgetting; this category also includes optimization based  methods that project the parameters of new tasks in orthogonal null spaces \cite{adamnscl, orthogonal_cl_2, orthogonal_cl_3, orthogonal_cl_4} in order to avoid catastrophic interference. The third category of methods includes \emph{(3) replay} which either directly stores training data from previous tasks in a memory buffer~\cite{mem_replay_cl_1, mem_replay_cl_2, mem_replay_cl_3, mem_replay_cl_4} or train a generative model in order to synthesize data from the same distribution as the actual training data~\cite{gan_cl_1, vae_cl_1}. This third category, in particular, has attracted a lot of attention in the literature with various generative models being applied to incremental learning including GANs, \cite{gan_cl_1, gan_cl_2, gan}, VAEs \cite{vae_cl_1, vae_cl_2, vae}, normalizing flows (NFs) \cite{nf_cl_2, nf_citation}, and diffusion models (DMs) \cite{dm_cl_1, dm_cl_2, dm_og}. Nonetheless, these incrementally learned generative  models are known to be computationally demanding with a significant increase in time and memory footprint as the number of tasks increases. Some of those methods \cite{dm_cl_features, gan_feature_replay} mitigate the computational complexity of these models by replaying feature vectors instead of images. \\
\indent Most of the aforementioned generative methods consider (1) \textcolor{black}{class/task} dependent generative models~\cite{dm_cl_features, cgil}, or (2) heavily rely on self-replay \cite{gan_feature_replay, nf_cl_2, brain_inspired_replay_vae} in order to maintain the generalization power of the trained models on the previous classes; as a result, these two subfamilies of methods scale linearly in memory and time respectively. Ideally, a solution based on a unique static generative model, for all tasks, is preferred that would only be trained on data of the current task (and without self-replay) by considering gradient projection in the null space of the previous tasks. Note that amongst generative models, VAEs are the most suited for incremental learning as (i) GANs exhibit reduced diversity and unstable training \cite{gans_unstable}, (ii) NFs are constrained to be invertible and hence difficult (and also slow) to train, and less expressive for an equivalent number of parameters compared to other generative models \cite{papamakariosNormalizingFlowsProbabilistic2021c}, and (iii) DMs have long sampling times requiring several function evaluations on a neural network for long time-steps, making null space optimization highly non-trivial~\cite{continual_diffusion_anomaly}. In contrast, VAEs only require the decoder to be incremental, and usually have low dimensional latent spaces making gaussian parameter estimation much more tractable and more stable at frugal data regimes (as frequently observed in incremental learning). Furthermore, VAEs require only a single forward pass on lightweight networks for data generation.  \\
\noindent Considering all these issues, we propose to train replay-free incremental VAE which leverages optimization in the null space of the previous tasks. In order to facilitate the learning of new classes, a multi-gaussian conditioning scheme is designed, in the VAE latent space, using a fixed-point iteration framework. The proposed method shares some similarities with the works \cite{brain_inspired_replay_vae, closed_loop_transcription_incremental_vae}, incrementally training VAEs for generative replay. However it significantly differs since those methods only train/estimate gaussians using a likelihood loss (no KLD repulsion), resulting in overlapping gaussians with erroneous labels that should  be corrected using a separate classifier network \cite{brain_inspired_replay_vae}.  Additionally, those methods suffer from extensive forgetting, since there is no null space optimization but only suboptimal self replay. Those differences also apply to \cite{nf_cl_2}, which uses an incremental NF with randomly initialized gaussian means (without KLD and null space optimization). \\
\textbf{Foundation Models \& Incremental Design.} Incremental learning strategies based on foundation models, generally avoid the costly update of large attention  models using efficient strategies. Considering CLIP, this translates into freezing image and text encoders, and adding dynamic (optionally conditional) parameters that are trained to optimize CLIPs performance towards a desired task. This includes prompt learning \cite{coop, cocoop, dualprompt,  codaprompt, attriclip, vpt, star-prompt, cgil, vpt_ns, pgp_prompt} which enriches the input given to a text encoder, in order to align it with image features obtained by an image encoder. Those prompts, however, may interfere with previous tasks \cite{star-prompt, consistent_prompt} which causes performance degradation. A solution mitigating this issue, consists in adding an orthogonality constraint on prompts \cite{codaprompt, dualprompt, attriclip, pgp_prompt, vpt_ns}. Other solutions based on generative feature replay obtain better performances by using a different variational autoencoder for each class as in CGIL \cite{cgil}. However, the drawback of this method resides in the large memory footprint when storing the generative models. Indeed, and as will be shown in the experiment section (in the lower part of Table~\ref{tab:ablation_second}), it is in fact preferable both in terms of memory and accuracy to directly store all extracted features for replay, instead of using a generative model. As such the proposed method in CGIL \cite{cgil} sidesteps the incremental learning challenge by using excessive amounts of memory to store all generators. Hence, to mitigate those issues, we build upon the above works and propose to train \textbf{a single incremental conditional VAE}, while being memory efficient, replay-free and ultimately reaching forgetting-free models overtaking these closely related works.

\section{Background}
Prior to introduce our main contribution in section~\ref{sec:method}, we briefly revisit in this section all the background in variational autoencoders and the used foundation models, namely CLIP.  
\subsection{Variational Autoencoders (VAEs)}
Variational Autoencoders (VAEs) [19] are likelihood-based generative models that maximize the evidence lower bound (ELBO), generally decomposed into a reconstruction loss and a prior matching loss  \cite{unified_perspetive_dm} as
\begin{equation}
	\label{eq:elbo}
	\mathcal{L}_{ELBO} = \underbrace{\mathbb{E}_{q_{{\phi}}(\z|\x)} [\log p_{\theta}(\x | \z)]}_{\text{reconstruction loss}} - \underbrace{D_{KL}(q_{\phi}(\z| \x)|| p(\z))}_{\text{prior matching loss}}.
\end{equation}
Here $q_{{\phi}}(\z|\x)$ is the encoder model parameterized by $\phi$ that learns to compress input data $\x$ into low dimensional gaussian parameters $\mubold_{\phi, \x}$ and $\sigmabold_{\phi, \x}$. The latter are optimized to be as close as possible to a prior $p(\z)$ (usually a standard gaussian $\mathcal{N}(\bf{0}, \bf{I})$) using KLD. The decoder $p_{\theta}(\x | \z)$  parameterized by $\theta$, learns to reconstruct input data $\x$ based on sampled latent variables $\z \sim \mathcal{N}(\mubold_{\phi, \x}, \sigmabold_{\phi, \x})$ using the reconstruction loss. In order to facilitate back-propagation, $\z$ is reparametrized as
\begin{equation}
	\label{eq:reparam}
	\z = \mubold_{\phi, \x} + \mathbf{\sigmabold_{\phi, \x}} \odot \bf{\epsilonbold},
\end{equation}
with $\epsilonbold \sim \mathcal{N}(\bf{0}, \bf{I})$ being a random variable.
\subsection{CLIP} \label{sec:clip_section}
Contrastive Language-Image Pre-Training (CLIP) \cite{clip} is an encoder based foundation model capable of zero-shot classification. It simultaneously trains an image encoder $E_{img}(.)$ and a text encoder $E_{txt}(.)$ to map matching images and textual descriptions to similar locations in the feature space. A function $\promptfn(\y)$ is generally used to construct a textual prompt given a label $\y$ such that $\promptfn(\y_{i}) = \emph{"A PHOTO OF [$\y_{i}$]"}$, where ($\x_{i}, \y_{i}$) are image-label pairs. That means the encoded variables $\z_{i}^{img} =  E_{img}(\x_{i})$ and $\z^{txt}_{i} = E_{txt}(\promptfn(\y_{i}))$ should be as close as possible in the \emph{feature space}. When fine-tuning a CLIP based model for classification, this is ensured by maximizing the probability of predicting the label $\y_{i}$ associated to $\x_{i}$ as
\begin{equation}
	\label{eq:cosine_similarity}
	p(\y_{i} |\x_{i}) = \frac{\exp(S_{C}(\z_{i}^{img}, \z_{i}^{txt})/\beta)}{\sum_{k=1}^{K} \exp(S_{C}(\z_{i}^{img}, E_{txt}(\promptfn(k)))/\beta)},
\end{equation}
where $S_{C}(.,.)$ is the cosine similarity, $\beta$ is a temperate scaling, and $K$ is the total number of classes. As showcased in~\cite{coop, cocoop, cgil}, when parametrizing the prompting function $\promptfn_{\omega}$(.), with parameters $\omega$, a  drastic improvement in classification tasks can be achieved.

\section{Proposed Method} \label{sec:method}
\paragraph{Problem Formulation.}
The core of our approach lies in building an incremental memory buffer using a lightweight generative model—specifically, a VAE decoder—which is then used to adapt a classifier network \textcolor{black}{on new tasks using synthetic samples}. More formally, given a stream of data divided into $T$ distinct groups called tasks $\{{\cal T}^t\}_{t=1}^T$, where a task ${\cal T}^t$ can be seen as a dataset  $\D\t = \{(\x^{t}_{i}, \y^{t}_{i})\}_{i}$ made of image-label pairs with a label space $\in \Y^{t}$. We seek to incrementally train a conditional VAE model $\incrmodel_{\incrmodelparams}(.)$ (parameterized by $\incrmodelparams$) to learn new classes without forgetting previous ones. Since labels are not overlapping between successive tasks: $\Y\t \cap \Y^{t'} = \emptyset $ for  $\forall t \neq t'$, our incremental learning consists in updating the parameters $\incrmodelparams$ on a current task $t$, while minimizing catastrophic forgetting on the set of classes from previous tasks denoted as $\Y^{1:t-1} =  \cup_{t'=1}^{t-1} \Y^{t'}$. \textcolor{black}{Afterwards, this novel incremental VAE, leveraging the proposed powerful incremental multi-gaussian conditioning, and greatly benefiting from orthogonal training, is used \textbf{to adapt a classifier network} --- as in a multi-task setting, by leveraging synthetic data generated by the decoder $p_{\theta}$.}  

\subsection{Incremental Setup}
\textcolor{black}{During the previously mentioned adaptation step, the CLIP classifier optimizes its class-incremental prompting parameters $\omega$ via Eq.~\ref{eq:cosine_similarity}, by feeding them to the text encoder $E_{txt}$, and leveraging synthetic data from all tasks generated by the decoder network $p_{\theta}(.)$.} Then, the adapted CLIP's classification accuracy is used as a proxy metric for assessing forgetting in the incrementally trained VAE $\incrmodel_{\incrmodelparams}$. During this prompt-learning process, the image $E_{img}$ and text $E_{txt}$ encoders are frozen, and solely the class-conditional prompts $\promptfn_{\omega}(\y)$ are trained. In this work, prompts are parameterized following the CGIL framework~\cite{cgil}, incorporating three components: (i) a class-conditional embedding inspired by CoOp~\cite{coop}, denoted as $\promptfn_{\omega_{1}}(\y)$; (ii) a class-conditional embedding learned via a shared MLP across all classes, similarly to CoCoOp~\cite{cocoop} and denoted as $\promptfn_{\omega_{2}}(\y)$; and (iii) the standard prompting function $\promptfn(\y)$ defined in  \ref{sec:clip_section}. The final prompt $\promptfn_{\omega}(\y)$, obtained as the concatenation of the prompts $\promptfn_{\omega_{1}}(\y)$, $\promptfn_{\omega_{2}}(\y)$, $\promptfn(\y)$, is then fed to the text encoder so that the textual representation $\z^{txt}_{\y} = E_{txt} (\promptfn_{\omega}(\y))$ is optimized  via Eq.~\ref{eq:cosine_similarity} to match conditional synthetic representation $\z_{\y}^{img}$ from the decoder. As such, this process severely relies on quality and reliability of data and labels generated by the decoder $p_{\theta}(.)$, \textcolor{black}{whose training process is significantly enhanced as described in the following sections.}
\subsubsection{Conditional Priors}
\begin{figure}[!ht]
	\centering
	\begin{subfigure}{0.22\columnwidth}
		\centering
		\includegraphics[width=1.0\columnwidth]{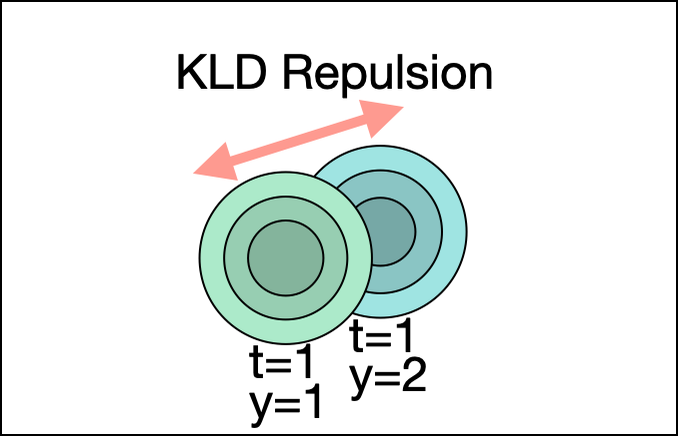}
		\caption{Task 1: initial \\ gaussians.}
		\label{fig:task1_init}
	\end{subfigure}%
	\begin{subfigure}{0.22\columnwidth}
		\centering
		\includegraphics[width=1.0\columnwidth]{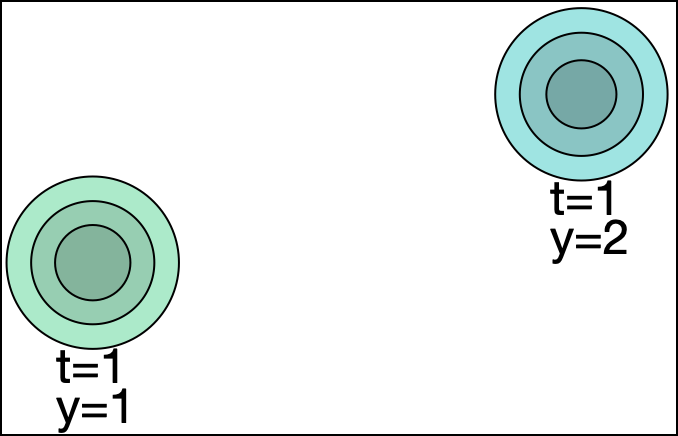}
		\caption{Task 1: learned \\ gaussians.}
		\label{fig:task1_final}
	\end{subfigure}
	\begin{subfigure}{0.22\columnwidth}
		\centering
		\includegraphics[width=1.0\columnwidth]{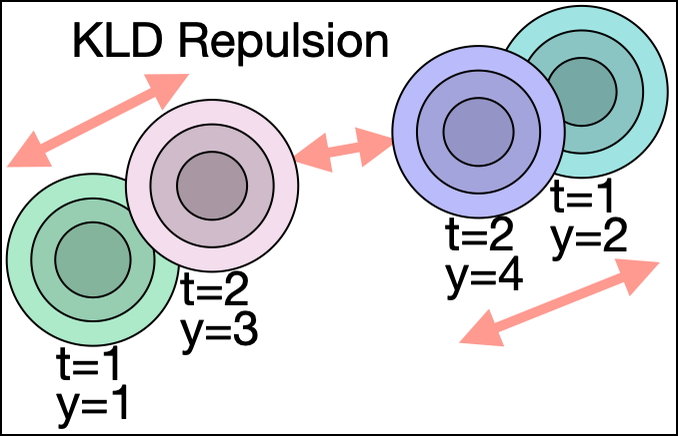}
		\caption{Task 2: initial \\ gaussians.}
		\label{fig:task2_init}
	\end{subfigure}%
	\begin{subfigure}{0.22\columnwidth}
		\centering
		\includegraphics[width=1.0\columnwidth]{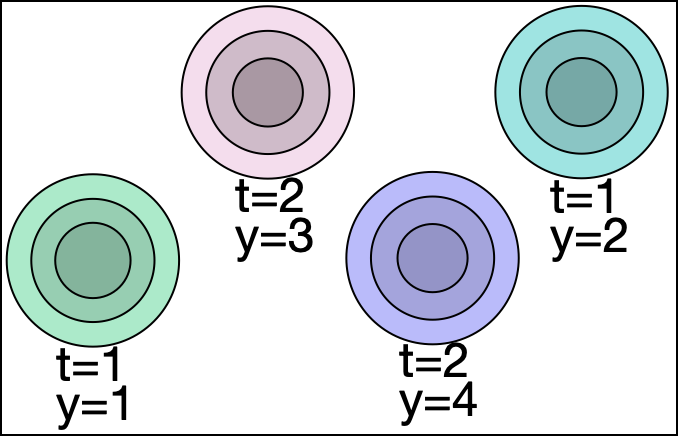}
		\caption{Task 2: learned \\ gaussians.}
		\label{fig:task2_final}
	\end{subfigure}
	\caption{Visualization of the incremental learning of gaussians using the fixed-point iteration. The variable \emph{'t'} denotes the index of a task, and \emph{'y'} the label of a class. Fig.~\ref{fig:task1_init} and ~\ref{fig:task2_init} respectively show the initial gaussians for task 1 and 2, corresponding to the conditional mean of data in the latent space, whereas Fig.~\ref{fig:task1_final} and ~\ref{fig:task2_final} respectively show the learned gaussians after the fixed-point optimization.}
	\label{fig:fpi_for_task_vis}
\end{figure}

To make the VAE conditional, the prior $p(\z)$ from Eq.~\ref{eq:elbo} is replaced with a conditional one $p(\z | \y)$, so that each class has a distinct gaussian in the latent space.  To learn the means $\mu_{\y}$ of each gaussian $\mathcal{N}_{\y}$ (covariances are kept fixed to the identity $\bf I$), a fixed-point iteration (FPI) is used as shown in Eq.~\ref{eq:fpi}, which is derived from the optimality conditions of Eq.~\ref{eq:fpi_full} (a proof is available in Section \ref{seq:fpi_proof} of the Appendix). This formulation is adopted from the work in \cite{nfpi},  which uses a fixed point iteration based on a likelihood, and a KLD loss applied between gaussian pairs in order to learn a multimodal latent space. However it differs on several aspects from our work, most notably (i), since it is not applied to continual learning nor VAEs, but to generative classification and image augmentation using an NF model only. Secondly, (ii) the equations involve full covariance matrices, and their estimation is not feasible in incremental learning, especially when limited amounts of data are available for training. As such, we only consider simpler isotropic covariance matrices, that also do not result in an extra memory cost. Finally, (iii) we derive a more interpretable equation involving only a squared L2 norm between gaussian means, by applying a negative log on the KLD pairs (instead of the exponential), which is very intuitive and provides an explanation on how gaussians are being separated.
\begin{equation}
		\mathcal{L}_{optim} =   \sum_{\substack{\x,\y \in \mathcal{D}^{t}}} -\log (\mathcal{N}(\mu_{\phi, \x, \y}; {\mu_{\y}, \bf{I}})) 
		+ \lambda \mysumyy-\log (D_{KD}(\mathcal{N}_\y \ || \  \mathcal{N}_{\y'}))
			\label{eq:fpi_full}
\end{equation}
\begin{equation}
	\label{eq:fpi}
	\mu_{\y}^{(\tau)} = \frac{1}{N_{\y}} \sum_{i=1}^{N_{\y}} \mu_{\phi, \x\i, \y\i }  + \lambda \mysumy \frac{\mu_{\y}^{(\tau-1)} - \mu_{\y'}^{(\tau - 1)}} {\big \lVert \mu_{\y}^{(\tau-1)} - \mu_{\y'}^{(\tau - 1)} \big \rVert_{2}^{2}}.
\end{equation}
Here  in Eq.~\ref{eq:fpi}, $\mu_{\y}^{(\tau)}$ refers to the mean obtained at iteration $(\tau)$ as a function of the means at iteration $(\tau-1)$. In Eq.~\ref{eq:fpi_full}, the first term corresponds to negative log-likelihood that maximizes the probability of the latent data $\mu_{\phi, \x, \y}$ belonging to a conditional gaussian, and the second term is the negative log of the pairwise Kullback-Leibler divergences applied to gaussian pairs to repel them apart. The use of the negative log ensures both terms are homogeneous and balanced, and that the second term is \emph{"virtually"} lower bounded, for high values of KLD.  The left-hand side term  of Eq.~\ref{eq:fpi} is the  conditional mean of the data in the latent space of the VAE, \textcolor{black}{where $N_{y}$ denotes the number of images with label $\y$ ie., $\y_{i} = \y$}. The right-hand side term acts as a repulsion criterion that seeks to amplify, proportionally to the  setting of $\lambda$, the distances between means belonging to different classes. Overall, the interpretation of Eq. \ref{eq:fpi}  is very intuitive and one may recognize the impact of the KLD term. \noindent In an incremental setting, note that means belonging to the previous tasks (and the underlying classes) are used when evaluating Eq.~\ref{eq:fpi} but remain fixed in order to prevent forgetting. Figure~\ref{fig:fpi_for_task_vis} illustrates this incremental learning process. At the beginning of a task, the latent means evaluated, after forwarding the data with the encoder, are overlapping (see Fig.~\ref{fig:task1_init}) and unreliable for conditional image generation. After applying the fixed-point iteration,  separated gaussians are obtained as in Fig.~\ref{fig:task1_final}, which are used to train the encoder and decoder following Eq.~\ref{eq:elbo_conditional}. Given a new task (see Fig.~\ref{fig:task2_init}), means are overlapping again, and by applying the fixed-point iteration algorithm on the new means, the latter are updated to avoid overlapping (as shown in Fig.~\ref{fig:task2_final}).

\subsubsection{Total loss of the VAE}
By taking into account all the conditioning added to the VAE, the global loss written in Eq.~\ref{eq:elbo_conditional} is minimized using only data $\mathcal{D}^{t}$ of the current task $t$. In the remainder of this paper (unless explicitly mentioned) we omit $t$ from the definition of all variables, so that the total loss can be written as 
\begin{equation}
		\mathcal{L}_{VAE} = \mathbb{E}_{q_{{\phi}}(\z|\x, \y)} [\log p_{\theta}(\x | \z, \y)]
		- D_{KL}(\mathcal{N}(\mubold_{\phi, \x, \y}, \sigmabold_{\phi, \x, \y})|| \mathcal{N}(\mu_{\y}, \bf{I})).
		\label{eq:elbo_conditional}
\end{equation}
\subsection{Orthogonality}
When ``naively'' optimizing the parameters of the decoder using Eq.~\ref{eq:elbo_conditional}, previous classes could be forgotten by \textcolor{black}{the learned decoder $p_\theta$}. In order to avoid this issue, and following~\cite{adamnscl}, the gradients $\{g_{l}\}_l$ of the weight parameters, through different layers of \textcolor{black}{$p_\theta$ in the current task}, are projected in the approximate null space of all previous tasks using an orthogonal projector $\pi_{l}$~\cite{maala}, i.e.,
\begin{equation}
	\label{eq:orthogonal_gradients}
	\overline{g_{l}} = \pi_{l}(g_{l}), 
\end{equation}
being
\begin{equation}
	\label{eq:projector}
	\pi_{l}(v) = U^{B}_{l} \cdot (U^{B}_{l})^{\top} \cdot v,
\end{equation}
and 
\begin{equation}
	\label{eq:svd_decomposition}
	U_{l} \cdot \Lambda_{l} \cdot U_{l}^\top = \Sigma_{l} \; \ \ \ \text{where}\; \ \ \  U_{l} =  (U^{A}_{l},  U^{B}_{l}).
\end{equation}
\noindent The projector $\pi_{l}$ in Eq.~\ref{eq:projector} is defined using the eigenvectors $U^{B}_{l}$ associated to the $|B|$-smallest eigenvalues of the layer-wise covariance matrix $\Sigma_{l}$ of data belonging to  $\Y^{1:t-1}$.  Note that despite being incrementally estimated for each task, the covariance matrix $\Sigma_{l}$ is exact. Besides, this orthogonal gradient guarantees that parameter updates will not affect the previous class distributions modeled \textcolor{black}{by $p_\theta$} while being able to learn new task distributions; in other words, it allows handling catastrophic forgetting while learning new tasks effectively (as shown later in experiments).

\subsection{Why forgetting is mitigated?}
This section outlines the benefit of the joint interaction of our multi-modal latent space design and null-space optimization in mitigating catastrophic forgetting within our incremental VAE. Given an incremental decoder $p_{\theta_{t}}$ with parameters  $\theta_{t}$ optimized on the current task $\mathcal{T}^{t}$, the core challenge lies in preserving its ability to generate reliable synthetic samples and labels for all preceding tasks $\mathcal{T}^{1:t-1}$. This objective is primarily bolstered by the incremental multi-modal latent space, characterized by well-separated gaussians (see the separability in Fig~\ref{fig:tsne_fpi_incremental_visualization} of the Appendix). The distinct gaussian priors in the latent space significantly streamline the conditioning process for our VAE, minimizing confusion compared to a standard normal gaussian $\mathcal{N}(\mathbf{0}, \mathbf{I})$ (as empirically corroborated in Table  \ref{tab:ablation}). Secondly, the optimization in the null-space ensures the immutability of gaussians associated with prior tasks $\mathcal{T}^{1:t-1}$ as the decoder exhibits negligible forgetting. Consequently, the features generated by $p_{\theta_{t-1}}$ and  $p_{\theta_{t}}$ on tasks $\mathcal{T}^{1:t-1}$ remain virtually indistinguishable, a result corroborated by Table~\ref{table:ratio_all_layers}  in the Appendix, which demonstrates a minimal ratio between the summed eigenvalues $U^{B}_{l}$ and $U_{l}$. Besides, gaussian separation achieved through FPI conditioning enhances further null-space optimization, preventing the decoder from receiving overlapping or similar input samples in subsequent tasks, as it happens with a vanilla prior  $\mathcal{N}(\mathbf{0}, \mathbf{I})$ where all classes/tasks are mingled in a single gaussian. Hence, multi-modal gaussians prevent rapid saturation of the null space (as later shown in Table \ref{table:contact_layer_orthogonality} of the Appendix). Finally, the non-incremental training of the encoder $q_{\phi_{t}}$ enables a more optimal minimization of the prior matching loss during our VAE's incremental learning. 
\subsection{Overcoming Dimensionality Bottleneck}\label{dyna}
As the gradients of the decoder, for a current task $t$, belong to the null space of $\Y^{1:t-1}$ , this space could be exhausted resulting in a dimensionality bottleneck that may prevent the VAE model $h_\psi$ from learning new tasks. In order to circumvent this issue while also increasing the expressiveness of the decoder, we consider multiple subnetwork instances of $h_\psi$  to be task-dependent. In practice, these subnetwork instances correspond to the last fully connected layer as well as the biases of the decoder; in total, only a small fraction of the parameters of $h_\psi$ \textcolor{black}{grows linearly w.r.t. the number of tasks}, in order to mitigate dimensionality bottleneck and allow learning new tasks more effectively as shown later in experiments. Algorithm \ref{alg1} available in the Appendix further details the incremental learning framework; in sum, the proposed method starts by converting an image dataset from task $\Tau^{t}$ to a feature dataset $\D\t$ using the CLIP image encoder. Then, multi-gaussian conditioning is learned using the fixed-point iteration (as shown in Eq.~\ref{eq:fpi}), prior to training both the encoder and decoder (as shown in Eq.~\ref{eq:elbo_conditional}). Afterwards, the eigenvectors of the null space of $\Y^{1:t-1}$  are updated in order to take into account the current task $t$. Finally, a synthetic dataset is generated  using the incrementally learned and conditioned decoder; this dataset is used to adapt the prompting parameters $\omega$ of the CLIP classifier to all classes in  $\Y^{1:t}$ (using Eq.~\ref{eq:cosine_similarity}). In Figure \ref{fig:pipeline} of the Appendix, we also illustrate the generative framework, including conditional priors, embeddings, orthogonality, and parameter instances in order to overcome dimensionality bottleneck.

\section{Experiments} \label{sec:experiments}
This section shows the impact of the proposed incremental VAE on image classification datasets and its comparison against different baselines, as well as an ablation study that shows the impact of each component of our model. A comparison against the related works is also provided.
\subsection{Datasets and Evaluation Metrics}
\textbf{Datasets.} We evaluate the proposed method on a wide range of challenging datasets, commonly used in incremental learning with CLIP based PEFT methods.  Those include fine-grained classification datasets such as \emph{Split CUB-200 \cite{cub200}} containing 200 classes divided into 10 equal tasks, and \emph{Split CARS-196 \cite{cars196}} which comprises 196 car classes split into 10 tasks, with the final task containing 16 classes while other tasks contain 20 classes each. Additionally, we also consider general knowledge datasets such as \emph{Split ImageNet-R~\cite{imagenet-r} and Split CIFAR100 \cite{cifar}} which respectively contain 200 and 100 classes, split into 10 equally sized tasks. \\
\textbf{Metrics.} We evaluate the performances of the proposed method in class-incremental learning (CIL) scenario which reports the Final Average Accuracy (FAA) scores obtained on each task, after encountering the last task. In all experiments, the results are averaged over 3 seeds, and we report the means and the standard deviations.

\subsection{Baselines \& Methods}
In order to compare our model, we consider different baselines in our experiments. \\ 
\noindent \textbf{Baseline 1: Unconditional VAE (VAE-CGIL).} This baseline --- identical to CGIL \cite{cgil} --- trains an unconditional VAE for each class, and stores the decoder in a memory buffer. All proposed methods build upon this one by training \textbf{a single incremental VAE}. \\
\noindent \textbf{Baseline 2: Conditional Embedding (VAE-CEO).} This baseline learns a separate conditional embedding concatenated to each fully connected layer of the encoder and decoder. Note that this baseline does not rely on multiple gaussians in the latent space, but instead it uses a single gaussian $\mathcal{N}(\bf{0}, \bf{I})$. In this baseline, the decoder is trained with orthogonality. \\
\noindent \textbf{Our method (variant 1: VAE-FO).} This proposed method relies on the FPI for conditioning, and trains a static decoder with orthogonality. \\
\noindent \textbf{Our method (variant 2: VAE-FOD).} This method builds upon VAE-FO architecture, instantiating different last fully connected layers and biases, for each task in the decoder. \\
\noindent \textbf{Our method (variant 3: VAE-FODCE).} This method builds upon VAE-FOD, and adds a conditional embedding as in the baseline VAE-CEO for a more expressive double conditioning.  

\noindent \textbf{Implementation Details:} The encoder and decoder are made of 3 fully connected layers, with latent and hidden layers including 256 and 512 dimensions respectively, the dimension of the first (resp. last) fully connected layers of the encoder (resp. decoder) is 768 corresponding to CLIP features. At each task, the VAE is trained for 500 epochs using an Adam optimizer with a learning rate set to $5.0^{-4}$; this learning rate is reset to  $5.0^{-5}$  when orthogonality is used. The $\lambda$ factor in Eq.~\ref{eq:fpi} is set to 900 for all experiments, and the fixed-point iteration process is run once (till convergence) at the beginning of each task, \textcolor{black}{since very distinctive separability is achieved as shown in Figure~\ref{fig:tsne_fpi_incremental_visualization} of the Appendix.}  Regarding the null space optimization, the eigenvectors  $U_{l}^{B}$ kept for  projection (with Eqs.~\ref{eq:projector}~and~\ref{eq:svd_decomposition}) correspond to the eigenvalues smaller than $a \cdot \lambda_{l}^{min}$, being  $\lambda_{l}^{min}$ the smallest eigenvalue, and $a$ set to $100$ in all experiments and for all datasets, including those without CLIP (as shown in Section \ref{seq:results_without_clip} of the Appendix). The dimension of the conditional embedding added to all layers of the underlying baseline is 10. As in \cite{attriclip, cgil, star-prompt}, we use a CLIP model based on ViT-L/14 and we follow the training procedure in \cite{cgil} for fair comparisons. 

\subsection{Results} 
\noindent \textbf{Ablation Study.} Table~\ref{tab:ablation} shows an ablation study of our model with different settings including  FPI and orthogonality when taken separately and jointly (i.e., VAE-CEO vs. VAE-FO); this ablation is also shown with and without frozen decoders. Note that with orthogonality, the decoder $p_{\theta}$ is necessarily retrained, whilst without orthogonality the decoder could either be retrained or frozen \textcolor{black}{so as to ensure it does not suffer from forgetting}. From this table, we observe the positive impact of FPI (either with frozen or retrained decoders), and when combined with orthogonality, performances improve further. In other words, catastrophic forgetting is highly mitigated when both FPI (for class conditioning) and orthogonality (for parameter retraining) are leveraged. \textcolor{black}{Following \cite{dm_cl_features}, we also add an ablation on a classwise normalization before feeding the data to the encoder $q_{\phi}$, that we find beneficial for reducing the saturation rate of the null space optimization in the decoder $p_{\theta}$ (see Table \ref{table:contact_layer_orthogonality} in the Appendix for more details), which is also followed by the classwise de-normalization after generating a sample with the decoder $p_{\theta}$,  further refreshing the class conditional information of the samples.} Table~\ref{tab:ablation_second} (upper part) shows further analysis of the VAE-FO with the step described in section~\ref{dyna}, that relies on a dynamic architecture which overcomes dimensionality bottleneck, and also greatly increases the expressivity of the trained models (i.e., VAE-FOD). From these results, we observe a clear gain when combining these two steps. Finally, the combination of both conditioning methods (conditional gaussians and embeddings) in VAE-FODCE, reaches the highest scores on fine-grained classification tasks, indicating the importance of enhanced conditional information. Table ~\ref{tab:ablation_second} (lower part) shows a comparison of the proposed methods with the closely related work involving generative replay by training one unconditional VAE for each class (VAE-CGIL), and ``upper bound''  which stores all the historical data in a memory buffer. We observe that our method (VAE-FO) outperforms ``upper bound'' on 2 datasets namely CUB200 and ImageNet-R while respectively requiring 1.5 and 6.2 times less memory; when comparing against VAE-CGIL, our method requires 53 times less memory. This shows that FPI+orthogonality (even with static decoders) make it possible to better generalize, and allow producing better (diverse) data that further enhance the classification performance of CLIP. Note that the cost for learning a new class is even lower: 3.3k parameters for VAE-FO vs 788k for VAE-CGIL which corresponds to 238 times less parameters. Furthermore, it should be noted the non-incremental baseline, simply storing all data features consistently outperforms CGIL \cite{cgil} using up to 34 times less memory on CUB200, and this further justifies the necessity of our proposed incremental generative model.  Finally, Table \ref{tab:ablation_gaussians_mean} of the Appendix shows the enhanced benefit of the FPI conditioning, when compared with other conditional multi-gaussian based methods.  

\begin{table*}[!ht]
	\caption{{Ablation setting for the conditional mechanism and the null space optimization. FPI conditioning uses the fixed-point iteration described in Eq.~\ref{eq:fpi} to learn class conditional means for the gaussians. When it is not used, conditional embeddings of dimension 10, are used with a standard gaussian $\mathcal{N}(\bf{0}, \mathbf{I})$ in the latent space. Frozen decoder means there is no orthogonality-based (null space) optimization, and that the decoder is frozen after training on the first task. It can be noted that the FPI conditioning consistently achieves the highest final accuracy scores on all datasets. The gain compared to the  conditional embedding baseline is the highest on fine-grained datasets such as CUB200 and Cars196, indicating FPI conditioning is less prone to labeling mistakes. Both methods have an equivalent number of parameters (in millions): 2.38M for VAE-FO and 2.40M for VAE-CEO, when doing null space optimization. As suggested in \cite{dm_cl_features}, \textcolor{black}{we also add a classwise normalization + de-normalization as a final ablation component.} }}
	\label{tab:ablation}
	\centering
	\resizebox{1.0\linewidth}{!}{%
		\begin{tabular}{cccccccc} 
			\toprule
			\multirow{2}{*}{Method} & \multicolumn{3}{c}{Ablation Setting} & \multicolumn{4}{c}{Dataset} \\
			\cmidrule(lr){2-4} \cmidrule(lr){5-8}
			& FPI conditioning & \begin{tabular}[c]{@{}c@{}}Null space \\ optimization \\ for the decoder\end{tabular} & \begin{tabular}[c]{@{}c@{}}Frozen decoder\\after first task\end{tabular} & CUB200 & Cars196 & Imagenet-R & CIFAR100 \\
			\midrule
			\multirow{2}{*}{VAE-CEO} & \multirow{2}{*}{\xmark} & \multirow{4}{*}{\xmark} & \xmark & $11.63_{\pm 0.21}$ & $14.21_{\pm 0.84}$ & $12.93_{\pm 0.37}$ & $12.10_{\pm 1.40}$ \\
			& & & \cmark & $10.69_{\pm 0.50}$ & $10.64_{\pm 0.33}$ & $10.76_{\pm 0.23}$ & $12.55_{\pm 0.51}$ \\
			\cmidrule(lr){1-2} \cmidrule(lr){4-8}
			\multirow{2}{*}{VAE-FO} & \multirow{2}{*}{\cmark} & & \xmark & $44.44_{\pm 0.64}$ & $44.25_{\pm 1.40}$ & $32.32_{\pm 0.42}$ & $28.34_{\pm 2.00}$ \\
			& & & \cmark & $57.71_{\pm 0.51}$ & $60.01_{\pm 3.47}$ & $62.41_{\pm 1.62}$ & $63.20_{\pm 2.02}$ \\
			\midrule
			VAE-CEO & \xmark & \multirow{2}{*}{\cmark} & \multirow{2}{*}{\xmark} & $20.68_{\pm 1.01}$ & $29.93_{\pm 1.40}$ & $63.23_{\pm 0.50}$ & $75.04_{\pm 0.36}$ \\
			VAE-FO & \cmark & & & $\mathbf{69.28_{\pm 0.14}}$ & $\mathbf{73.66_{\pm 0.52}}$ & $\mathbf{76.94_{\pm 1.33}}$ & $\mathbf{77.26_{\pm 0.56}}$ \\
			\midrule
			\begin{tabular}[c]{@{}c@{}}VAE-FO \\+ Classwise-norm \end{tabular} & \cmark & \cmark & \xmark & $\mathbf{83.70_{\pm 0.23}}$ & $\mathbf{89.16_{\pm 0.23}}$ & $\mathbf{89.99_{\pm 0.08}}$ & $\mathbf{83.67_{\pm 0.25}}$ \\
			\bottomrule
		\end{tabular}
	}
\end{table*}

\begin{table}[!ht]
	\caption{Upper Part: Extra analysis of our model, with a ``dynamic architecture'' where a different last fully connected layer (and biases) is instantiated in the decoder for each task . The ``dynamic architecture'' setting obtains the best results, and when combining it with both conditioning methods (i.e., VAE-FODCE), highest improvements can be observed on fine-grained classification datasets. Lower Part: Comparison of the proposed method in terms of accuracy and memory cost with respect to (i) an upper bound storing all the encountered data for memory replay, (ii) a method training a single VAE for each class: VAE-CGIL \cite{cgil} that we retrain to be in the same conditions as our models.}
	\label{tab:ablation_second} 	
	\centering
	\resizebox{1.0\linewidth}{!}{%
	\begin{tabular}{ccccccccc} 
		\toprule
		\multirow{2}{*}{Ablation~Setting}                                                         & \multicolumn{2}{c}{CUB200}                                                              & \multicolumn{2}{c}{Cars196}                                                                               & \multicolumn{2}{c}{Img-R}                                             & \multicolumn{2}{c}{CIFAR100}                                  \\
		& Acc                                          & Mem (M)                                  & Acc                                                            & Mem (M)                                  & Acc                                          & Mem (M)                & Acc                                 & Mem (M)                 \\ 
		\midrule
		VAE-FO                                                                             & 83.70\textsubscript{±0.23}                   & \textbf{\textbf{\textbf{\textbf{2.96}}}} & 89.16\textsubscript{±0.23}                                    & \textbf{\textbf{\textbf{\textbf{2.96}}}} & 89.99\textsubscript{±0.08}                   & \textbf{\textbf{2.96}} & 83.67\textsubscript{±0.25}          & \textbf{\textbf{2.63}}  \\ \hline
		VAE-FOD  = VAE-FO \\+ Dynamic architecture                                                                 & 84.35\textsubscript{±0.45}                   & 6.26                                     & 90.02\textsubscript{±0.19}                                     & 6.26                                     & \textbf{\textbf{90.25\textsubscript{±0.16}}} & 6.26                   & 84.92\textsubscript{±0.14}          & 5.92                    \\ \hline
		VAE-FODCE = VAE-FO \\+ Dynamic architecture\\+ Conditional Embedding                                                                                  & \textbf{\textbf{84.51\textsubscript{±0.24}}} & 6.39                                     & \textbf{\textbf{\textbf{\textbf{90.61\textsubscript{±0.07}}}}} & 6.39                                     & 90.07\textsubscript{±0.10}                   & 6.39                   & 84.00\textsubscript{±0.37}          & 6.05                    \\ 
		\hline\hline
		Comparative Baselines                                                                     &                                              &                                          &                                                                &                                          &                                              &                        &                                     &                         \\ 
		\hline
		VAE-CGIL \cite{cgil}                                                                      & 83.17\textsubscript{±0.11}                   & 158.26                                   & 89.27\textsubscript{±0.14}                                     & 158.26                                   & 89.01\textsubscript{±0.31}                   & 158.26                 & 86.02\textsubscript{±0.06}          & 79.13                   \\ 
		\hline
		\begin{tabular}[c]{@{}c@{}}Upper-bound: store all \\ data in a memory buffer\end{tabular} & 83.54\textsubscript{±0.21}                   & 4.60                                     & 90.20\textsubscript{±0.08}                                     & 6.25                                     & 89.64\textsubscript{±0.15}                   & 18.43                  & \textbf{86.72\textsubscript{±0.21}} & 38.40                   \\
		\bottomrule
	\end{tabular}
	}
\end{table}

\begin{table*}[!ht]
	\caption{Comparison results against other SOTA methods on standard incremental benchmarks. NA stand for not available, and the memory is expressed in millions of parameters. We reports the results \cite{cgil, star-prompt, slca, slca++} following their respective papers, and the remaining from the table provided in \cite{star-prompt}. We also additionally re-run the methods in \cite{moe} and \cite{cgil} on CIFAR100 with the codebase provided in \cite{cgil}, since the results were missing. It can be noted that VAE-FODCE achieves a new SOTA result on Cars196, outperforming the previous best method (CGIL \cite{cgil}) by up to 1.3 accuracy points.}
	\label{table:resutls_comparison}
	\centering
	\resizebox{\linewidth}{!}{%
		\begin{tabular}{cccccccc} 
			\toprule
			\textbf{Method} & \textbf{CUB200} & \textbf{Cars196} & \textbf{Img-R} & \textbf{CIFAR100} & Strategy & Replay Type & \begin{tabular}[c]{@{}c@{}}Replay memory \\Size in M \\(for 200 classes)\end{tabular} \\
			\midrule
			L2P \cite{l2p} & 62.21\textsubscript{±1.92} & 38.18\textsubscript{±2.33} & 66.49\textsubscript{±0.40} & 82.76\textsubscript{±1.17} & \multirow{4}{*}{Prompt} & \multicolumn{2}{c}{\multirow{5}{*}{NA}} \\
			DualPrompt \cite{dualprompt} & 66.00\textsubscript{±0.57} & 40.14\textsubscript{±2.36} & 68.50\textsubscript{±0.52} & 85.56\textsubscript{±0.33} & & \multicolumn{2}{c}{} \\
			CODA-Prompt \cite{codaprompt} & 67.30\textsubscript{±3.19} & 31.99\textsubscript{±3.39} & 75.45\textsubscript{±0.56} & 86.25\textsubscript{±0.74} & & \multicolumn{2}{c}{} \\
			AttriCLIP \cite{attriclip} & 50.07\textsubscript{±1.37} & 70.98\textsubscript{±0.41} & 86.25\textsubscript{±0.75} & 81.40 & & \multicolumn{2}{c}{} \\
			MoE Adapters \cite{moe} & 64.98\textsubscript{±0.29} & 77.76\textsubscript{±1.02} & \textbf{90.67\textsubscript{±0.12}} & 85.33\textsubscript{±0.21} & Dynamic Architecture & \multicolumn{2}{c}{} \\
			\midrule
			SLCA \cite{slca} & 84.71\textsubscript{±0.40} & 67.73\textsubscript{±0.85} & 77.00\textsubscript{±0.33} & \textbf{91.53\textsubscript{±0.28}} & Fine-tune + Replay & \multirow{2}{*}{\begin{tabular}[c]{@{}c@{}}Full Gaussian \\Replay\end{tabular}} & \multirow{2}{*}{118.19} \\
			SLCA++ \cite{slca++} & \textbf{86.59\textsubscript{±0.29}} & 73.97\textsubscript{±0.22} & 78.09\textsubscript{±0.22} & 91.46\textsubscript{±0.18} & LORA + Replay & & \\
			\midrule
			STAR-Prompt \cite{star-prompt} & 84.10\textsubscript{±0.28} & 87.62\textsubscript{±0.20} & 89.83\textsubscript{±0.04} & 90.12\textsubscript{±0.32} & \multirow{5}{*}{Prompt + Replay} & Diagonal MoGs & 1.54 \\ \hline
			CGIL \cite{cgil} & 83.12\textsubscript{±0.10} & 89.27\textsubscript{±0.14} & 89.42\textsubscript{±0.12} & 86.02\textsubscript{±0.06} & & \multirow{4}{*}{\begin{tabular}[c]{@{}c@{}}VAE \\Generative \\replay\end{tabular}} & 158.26 \\
			\textbf{VAE-FO (ours)} & 83.70\textsubscript{±0.23} & 89.16\textsubscript{±0.23} & 89.99\textsubscript{±0.08} & 83.67\textsubscript{±0.25} & & & 2.96 \\
			\textbf{VAE-FOD (ours)} & 84.35\textsubscript{±0.45} & 90.02\textsubscript{±0.19} & 90.25\textsubscript{±0.16} & 84.92\textsubscript{±0.14} & & & 6.26 \\
			\textbf{VAE-FODCE (ours)} & 84.51\textsubscript{±0.24} & \textbf{90.61\textsubscript{±0.07}} & 90.07\textsubscript{±0.10} & 84.00\textsubscript{±0.37} & & & 6.39 \\
			\bottomrule
		\end{tabular}
	}
\end{table*}
\noindent\textbf{SOTA Comparison. } Finally, Table~\ref{table:resutls_comparison} shows a comparison of the proposed method against closely related works involving PEFT methods and CLIP. In this table, we report the training and replay strategies used, as well as the memory footprint  when applicable. In Table \ref{table:sup_comparison_table_backbone_cgilcond} of the Appendix, we also report the backbones used. We observe that prompting methods without replay \cite{l2p, dualprompt, codaprompt, attriclip} generally reach the lowest accuracy scores, indicating interference between the prompts as new tasks are trained, and the importance of the replay. Note that the most closely related work corresponds to the VAE-based deep generative method from CGIL which uses a single unconditional VAE per class. From these results, our method (VAE-FODCE) successfully outperforms the SOTA  on Cars196 by 1.3 accuracy points using 24 times less memory for generative replay. As for other datasets, only MoE Adapters \cite{moe} reaches the highest accuracy without replay (on ImageNet-R), and it is based on a dynamic architecture with mixture of experts. In this table, we also compare against SLCA and SLCA++ which combine replay with either fine-tuning or LORA adaptation, however, the use of gaussians estimated with full covariances results in a higher memory footprint than the proposed method. Finally, we compare against STAR-Prompt which uses a prompt-based technique with a mixture of (diagonal) gaussians for replay (with a very low memory cost), but with larger inference time, since it requires an extra ViT-B/16 backbone compared to our method (86M extra parameters for the backbone). Nonetheless, we still outperform this comparative work on 3 datasets with a memory footprint (for a new class on VAE-FO) equal to 3.3k parameters vs 7.68k for STAR-Prompt, and this is twice larger and this memory footprint grows more quickly as classes are incrementally learned. 
\section{Conclusion}
In this paper, we propose a novel conditional VAE for incremental learning which is replay-free, and capable of learning new classes without degrading its performances on previous ones. The core contributions include (i) the optimization of training parameters in the span of the null space of previous tasks, greatly benefiting from (ii), a fixed-point conditioning that incrementally learns multi-gaussians in the latent space of the VAE, and achieving very robust label conditioning. Extensive experiments, conducted on different benchmarks, showcased the capability of our model to reach competitive SOTA accuracy against the closely related works while being more memory efficient.

{
	\small
	\bibliographystyle{unsrt}
	\bibliography{refs/generative_models, refs/datasets, refs/maths, refs/general_dl, refs/cl, refs/fpi, refs/nf}
}


\pagebreak
\clearpage

\appendix

\large

\normalsize
\onecolumn

\newcommand\tsnegraphsize{0.7}

\clearpage
\pagebreak

\large
\noindent The numbering of sections/lines/references in this Appendix follows the submitted paper.

\section{Content of the Appendix}

\paragraph{Content}
\label{sm}

\begin{figure*}[!ht]
	\includegraphics[width=\textwidth]{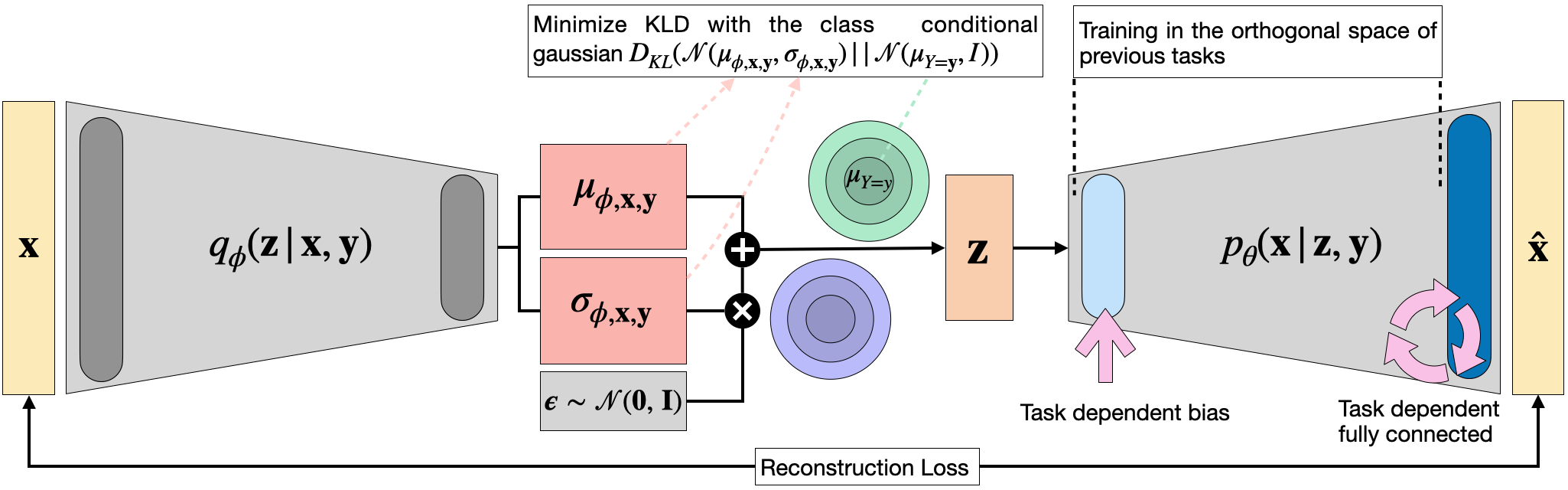}
	\caption{This figure shows the proposed architecture for incremental learning based on conditional VAE. The encoder $q_{\phi}(\z| \x, \y)$ and the decoder $p_{\theta}(\x| \z, \y)$ are both made of $L$ fully connected layers and are conditioned on priors $\mathcal{N}(\mu_{\y}, \bf{I})$ in the latent space. The decoder is trained by projecting gradients of new tasks in the null space of previous ones for all fully connected layers excepting the last one, which can optionally be task-dependent for more expressiveness, in which case, the biases are also made task-dependent as they correspond to very few parameters. The encoder on the other hand, is trained without any constraint on all tasks, and using the weights obtained at the end of a previous task. The variable $\hat{\x}$ generated by the decoder corresponds to the reconstructed $\x$ that was given as input to the encoder. } 
	\label{fig:pipeline}
\end{figure*}

\begin{itemize}
	\item Algorithm \ref{alg1} illustrating the incremental training pipeline. 
	
	\item Illustration of the proposed generative model in Figure~\ref{fig:pipeline}.
	
	\item Visualization of the accuracy at the end of each task is available in section \ref{seq:accuracy_at_each_task}.
	
	\item An analysis of the forgetting when using null space optimization is available in section \ref{seq:orthogonality implemetation details}.
	
	\item Comparison with other methods for the gaussian conditioning in the latent space in Table.~\ref{tab:ablation_gaussians_mean}.
	
	\item Backbones for Table \ref{table:resutls_comparison} of the main paper are available in Section~\ref{sup_sec:backbone_used_comparison_sota}.
	
	\item Results without a pretrained CLIP model, but a ResNet18 is available in section \ref{seq:results_without_clip}.

	\item Details regarding the computational time and ressources in section \ref{sec:computational_details_cgilcond}
	
	\item Discussion regarding the limitations of the proposed method in section \ref{sec:limitations_cgilcond}
	
	\item The derivation of the fixed point iteration formula from Eq.~\ref{eq:fpi} in the paper is available in  section \ref{seq:fpi_proof}.

\raggedbottom
\pagebreak
\clearpage

\end{itemize}	

\section{Algorithm of the incremental training pipeline}
\begin{algorithm}[]
	\footnotesize 
	\KwIn{Dataset ${\D_{img}^{t}}$ for $t \in [1, T]$  \tcp*[r]{\scriptsize Image dataset}} 
	Initialize the encoder $q_{\phi}(.)$ and the decoder $p_{\theta}(.)$ \\
	\For{$t \gets 1$ to $T$} {			
		$\D^{t}$  = $E_{img}({\D_{img}^{t}})$ \tcp*[r]{ \scriptsize Create feature dataset using CLIP image encoder}
		
		\For{$\mathbf{k} \in \Y\t$} {
			$\mu_{\mathbf{k}}$ = mean($\{\x_{i} | (\x_{i}, \y_{i}) \in \D\t , \y_{i} = \mathbf{k}\}$) \tcp*[r]{\scriptsize Calculate the class mean.} 
		}
		\vspace{0.1cm}
		$\tau \leftarrow 0$ \\
		$M^{(\tau)} = [\mu_{1}, ..., \mu_{|\Y^{1:t}|}] $  \tcp*[r]{\scriptsize Matrix of all means.} 
		
		\SetKwRepeat{Do}{do}{while}
		\Do{$\lVert M^{(\tau)} - M^{(\tau - 1)} \rVert_{2} > \epsilon $}{
			$M^{(\tau + 1)} = FPI(M^{(\tau)})$ 	\tcp*[r]{\scriptsize Update the means from current task using fixed-point iteration (FPI(.)) from Eq.~\ref{eq:fpi}.}
			$\tau \leftarrow \tau + 1$
		}
		\vspace{0.1cm}
		
		Train the incremental decoder $q_{\phi}(.)$ and encoder $p_{\theta}(.)$ on $\D\t$ following Eq.~\ref{eq:elbo_conditional}
		
		Update the eigenvectors  for each layer of the decoder following Eq. \ref{eq:projector} and ~\ref{eq:svd_decomposition}.

		$\D_{gen}^{t}  = \{\}$  \tcp*[r]{\scriptsize Synthetic dataset}
		\For{$\mathbf{k} \in \Y^{1:t}$} {
			\vspace{0.05cm}
			$\x_{\mathbf{k}}  = \text{decoder}(\z_{\mathbf{k}} ) | \z_{\mathbf{k}} \sim \mathcal{N}(\mu_{\mathbf{k}}, \bf{I})$ \tcp*[r]{\scriptsize generate artificial samples using the decoder} 
			$\D_{gen}^{t} \leftarrow \D_{gen}^{t}  \cup  \{ \x_{\mathbf{k}} , \mathbf{k}\}$  \tcp*[r]{\scriptsize add synthetic data to the synthetic dataset}
			
		}
		Adapt CLIP on $\D_{gen}^{t} $ to maximize the similarity using Eq.~\ref{eq:cosine_similarity}. \tcp*[r]{\scriptsize Learn the prompting parameters of CLIP text encoder.}
	}
	\caption{Incremental training pipeline}
	\label{alg1}
\end{algorithm}

\section{Illustration of the proposed generative model}

\pagebreak
\clearpage
\section{Visualization}
\subsection{Latent space}
\label{sec:latent_space_vis}
\begin{figure}[!h]
	\centering
	\begin{subfigure}{0.492\columnwidth}
		\centering
		\includegraphics[height=\tsnegraphsize\columnwidth]{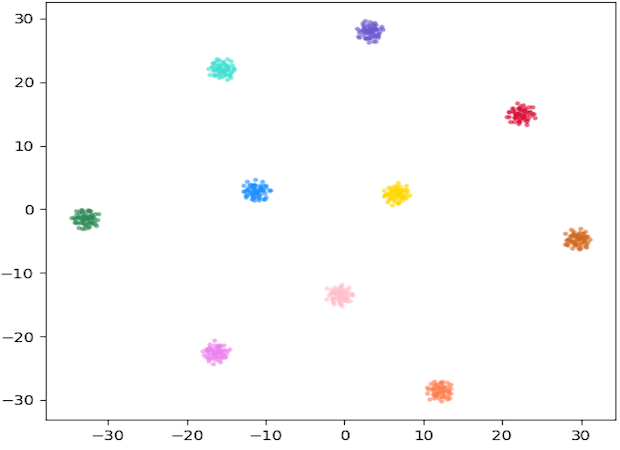}
		\caption{Task 1}
		\label{fig:task_1_tsne}
	\end{subfigure}%
	\begin{subfigure}{0.492\columnwidth}
		\centering
		\includegraphics[height=\tsnegraphsize\columnwidth]{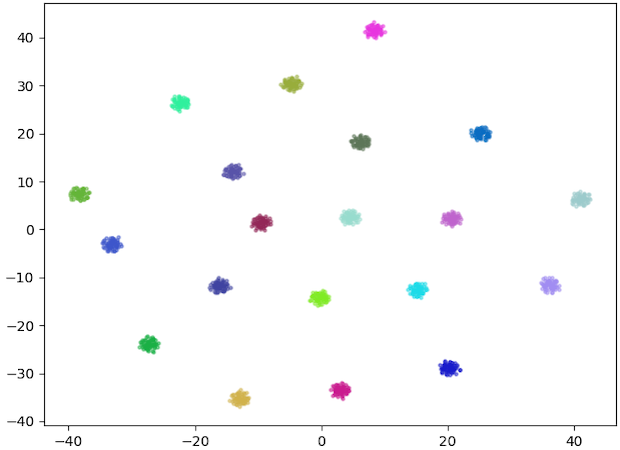}
		\caption{Task 2}
		\label{fig:task_2_tsne}
	\end{subfigure}%
	\quad
	\begin{subfigure}{0.492\columnwidth}
		\centering
		\includegraphics[height=\tsnegraphsize\columnwidth]{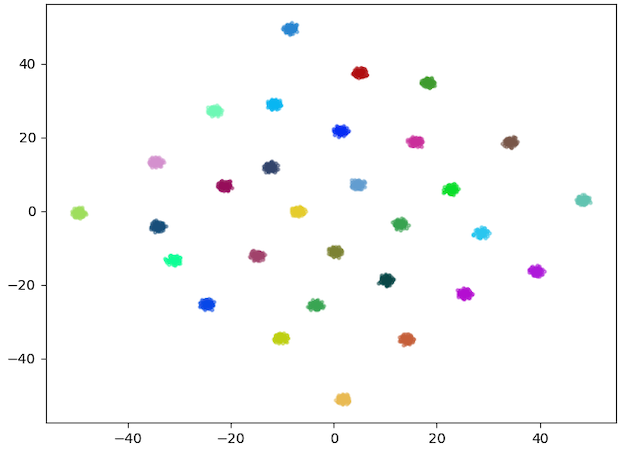}
		\caption{Task 3}
		\label{fig:task_3_tsne}
	\end{subfigure}%
	\begin{subfigure}{0.492\columnwidth}
		\centering
		\includegraphics[height=\tsnegraphsize\columnwidth]{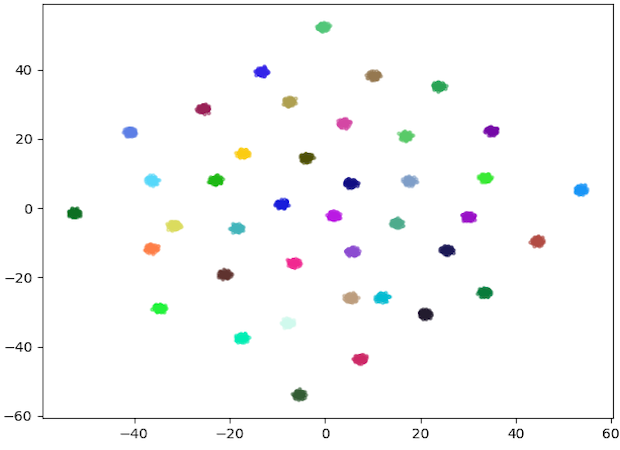}
		\caption{Task 4}
		\label{fig:task_4_tsne}
	\end{subfigure}
	\quad
	\begin{subfigure}{0.492\columnwidth}
		\centering
		\includegraphics[height=\tsnegraphsize\columnwidth]{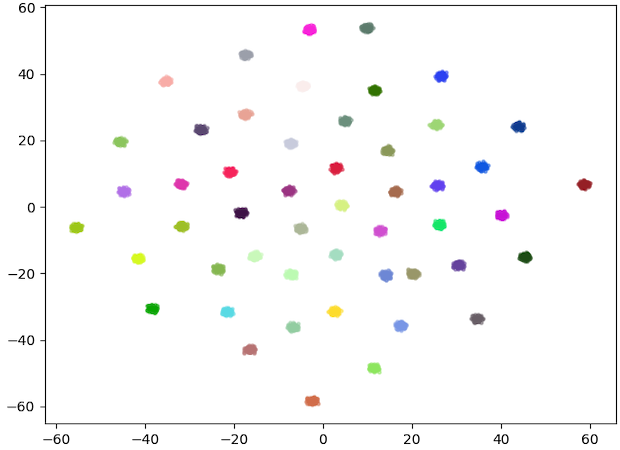}
		\caption{Task 5}
		\label{fig:task_5_tsne}
	\end{subfigure}%
	\begin{subfigure}{0.492\columnwidth}
		\centering
		\includegraphics[height=\tsnegraphsize\columnwidth]{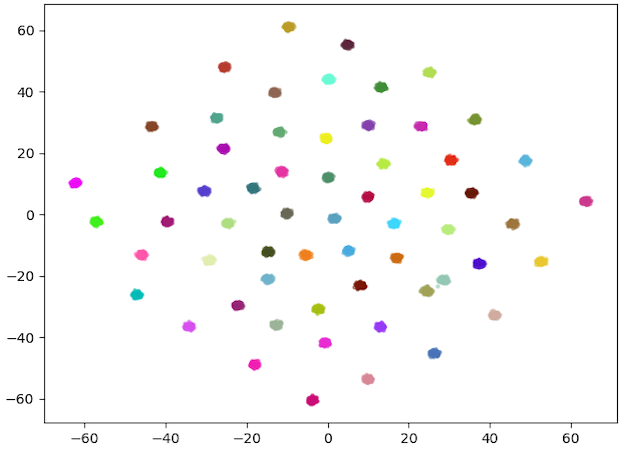}
		\caption{Task 6}
		\label{fig:task_6_tsne}
	\end{subfigure}%
	\caption{TSNE projection of classwise noise sampled from gaussians from task 1 to task 6 on CIFAR100 using VAE-FO. It can be noted that the gaussians incrementally learned using fixed-point iteration are clearly separated and non overlapping for all tasks.}
	\label{fig:tsne_fpi_incremental_visualization}
\end{figure}

\pagebreak
\clearpage
\subsection{Accuracy}
\label{seq:accuracy_at_each_task}

\newcommand\figsize{0.75}

\begin{figure*}[!ht]
	\center
	\includegraphics[width=\figsize\textwidth]{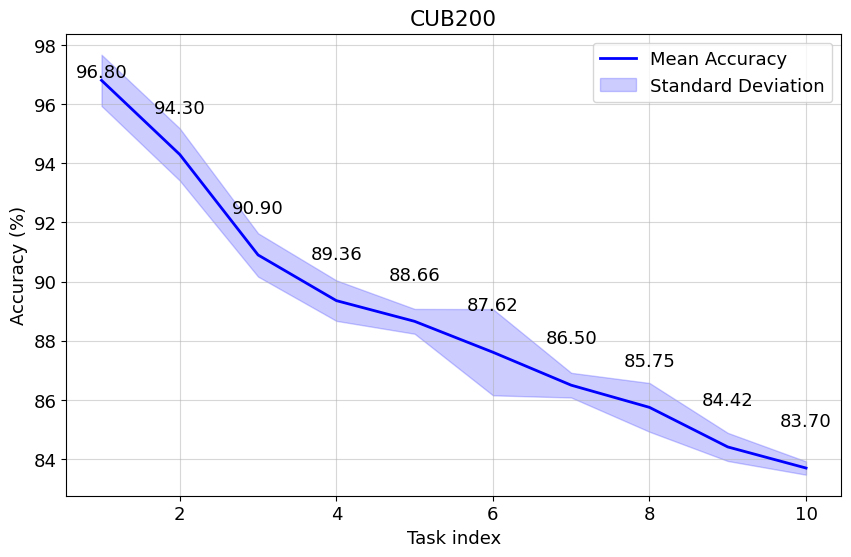}
	\caption{Accuracy and standard deviation on CUB200 as tasks evolve for VAE-FO.} 
	\label{fig:acc_cub}
\end{figure*}
\begin{figure*}[!ht]
	\center
	\includegraphics[width=\figsize\textwidth]{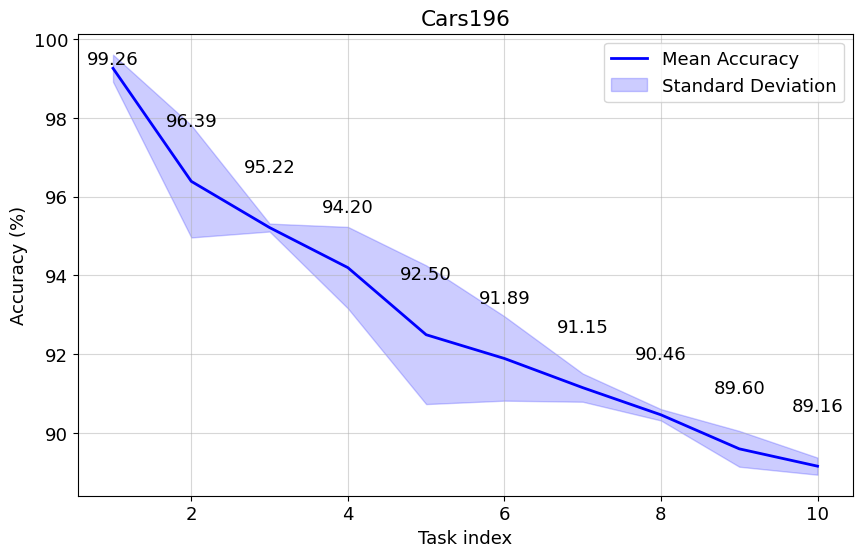}
	\caption{Accuracy and standard deviation on Cars196 as tasks evolve for VAE-FO.} 
	\label{fig:acc_cars}
\end{figure*}
\begin{figure*}[!ht]
	\center
	\includegraphics[width=\figsize\textwidth]{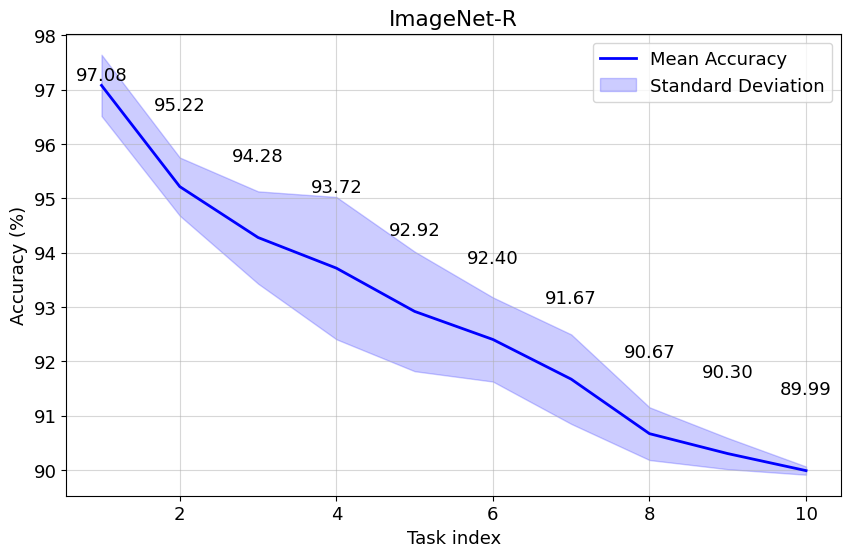}
	\caption{Accuracy and standard deviation on ImageNet-R as tasks evolve for VAE-FO.} 
	\label{fig:acc_imagenet-r}
\end{figure*}
\begin{figure*}[!ht]
	\center
	\includegraphics[width=\figsize\textwidth]{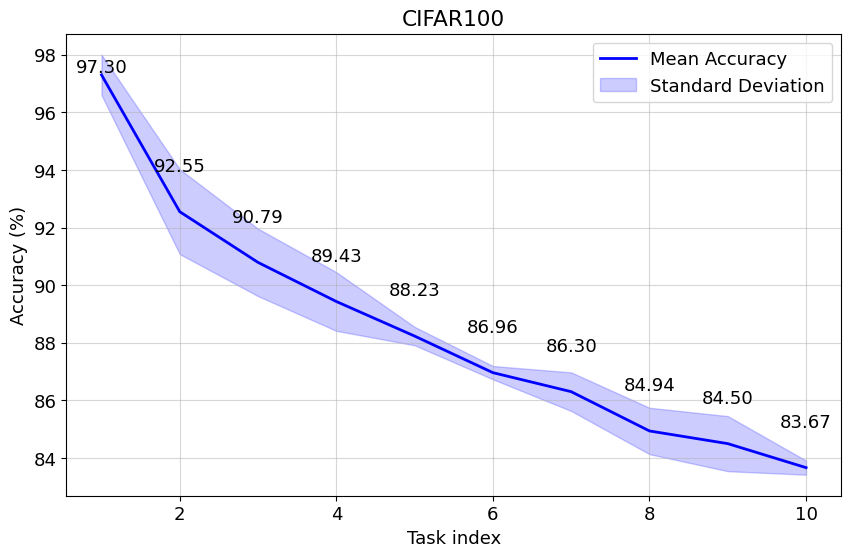}
	\caption{Accuracy and standard deviation on CIFAR100 as tasks evolve for VAE-FO.} 
	\label{fig:acc_cifar100}
\end{figure*}

\clearpage
\pagebreak

\section{Orthogonality}
\label{seq:orthogonality implemetation details}

In this section, we measure ratio  between the sum of the eigenvalues $\Lambda^{B}$ associated to $U_{l}^{B}$, and the eigenvalues $\Lambda$ associated to $U_{l}$ in the incremental decoder $p_{\theta_{t}}$. This ratio also called the \emph{"proportion"} and noted as $R$ following \cite{adamnscl}, can be calculated for all the layers that are trained in the null space. It is defined as:
\begin{equation}
	R = \frac{\sum_{\nu \in \Lambda^{B}} \nu}{\sum_{\nu \in \Lambda} \nu}
	\label{eqn:proportion}
\end{equation}
where $R \in [0, 1]$. \textcolor{black}{By multiplying this ratio $R$ by 100, one can think of it as the percentage of variance from previous tasks, that is present in the current approximate null space. As such, \textbf{If this ratio is very low \emph{i.e} close to 0, it indicates that parameter updates for the current task occur in the null space of all previous tasks, and this means that there is no forgetting. On the other hand, if this ratio is high \emph{i.e} close to 1, it indicates that the null space optimization has completely saturated, and that very significant forgetting (almost total forgetting) occurs on previous tasks, when optimizing the parameters for the current task.} As such, this ratio is a direct indicator of the catastrophic forgetting occurring in a layer / neural network optimized in the null space.} \\

\textcolor{black}{In table \ref{table:ratio_all_layers} we calculate the average proportion R  for all layers and all tasks in VAE-FO, and the very small result ($\leq 0.003$) indicates there is very little forgetting. In Table \ref{table:contact_layer_orthogonality} we investigate the effect of the classwise normalization on saturation rate of the null space, and also show that VAE-FO (multimodal latent space) has a lower forgetting according the proportion $R$ than VAE-CEO (unimodal latent space) in the first fully connected layer, which takes as input gaussian samples from the latent space. Indeed, this layer is the most subject to saturation/forgetting when using a unimodal gaussian $\mathcal{N}(\mathbf{0}, \mathbf{I})$, since samples are very likely to overlap in the latent space, and also because the dimension of this layer is the smallest.}

\begin{table}[!h]
	\caption{\textcolor{black}{Average proportion $R$ as defined in Eq.~\ref{eqn:proportion}, for all layers and all tasks in the decoder. This suggest the forgetting on CUB200 and Cars196 is infinitesimal on Imagenet-R and CIFAR100, since the ratio is respectively reaching 0.0027 and 0.027, which is smaller than 0.05 as originally set in \cite{adamnscl}}. Results in this table are reported for VAE-FO with classwise normalization.}
	\label{table:ratio_all_layers}
	\centering
	\resizebox{1.0\linewidth}{!}{%
	\begin{tabular}{lcccc} 
		\rowcolor[rgb]{1.0,1.0,1.0} 
		\toprule
		Dataset & CUB200 & Cars196 & Imagenet-R & CIFAR100 \\
		\midrule
		VAE-FO~ & $3.0 \times 10^{-8} \pm 8.1 \times 10^{-8}$ & $3.2 \times 10^{-8} \pm 4.5 \times 10^{-8}$ & $0.0027 \pm 0.0007$ & $0.027 \pm 0.01$ \\
		\bottomrule
	\end{tabular}
	}
\end{table}

\begin{table}[!h]
	\caption{Ablation on the impact of the classwise normalization for VAE-FO and VAE-CEO, by inspecting the average (on all tasks), of the proportion $R$ defined in Eq.~\ref{eqn:proportion} (the lower this value, the less forgetting occurs) in the first layer of the incremental decoder. This layer which is directly in contact with the multimodal latent space, is most subject to forgetting, and gives an indication on the importance of the structure of the latent space. Concerning the  classwise normalization, it can be noted that it greatly reduces the proportion ratio which mitigates forgetting in the decoder network. Furthermore, this ratio is much smaller for VAE-FO than for VAE-CEO,  respectively reaching 0.15 for VAE-FO vs 0.52 for VAE-CEO on CIFAR100 without the class normalization, which confirms the importance of the multimodal latent space.  It can also be noted that all values for VAE-FO (when doing class normalization) are very close to 0.05 which indicates very little forgetting is occurring. Furthermore this proportion ratio on CUB200 and Cars196 is infinitesimal with values close to $10^{-8}$, indicating there is virtually not forgetting (only numerical imprecisions).}
	\label{table:contact_layer_orthogonality}
	\centering
	\resizebox{1.0\linewidth}{!}{%
	\begin{tabular}{cccccc}
		\toprule
		Classwise Normalization & Method & CUB200 & Cars196 & Imagenet-R & CIFAR100 \\
		\midrule
		\multirow{2}{*}{\cmark} 
		& VAE-FO  & $5.9 \times 10^{-8}$   & $6.7 \times 10^{-8}$   & $\mathbf{0.052 \pm 0.01}$         & $\mathbf{0.069 \pm 0.027}$          \\
		& VAE-CEO & $4.2 \times 10^{-8}$   & $6.6 \times 10^{-8}$   & $0.059 \pm 0.01$                  & $0.140 \pm 0.022$                   \\
		\hline
		\multirow{2}{*}{\xmark} 
		& VAE-FO  & $1.3 \times 10^{-8}$ & $2.1 \times 10^{-8}$ & $\mathbf{0.09 \pm 0.16}$ & $\mathbf{0.15 \pm 0.01}$  \\
		& VAE-CEO & $1.7 \times 10^{-8}$ & $3.0 \times 10^{-8}$ & $0.15 \pm 0.25$          & $0.52 \pm 0.35$           \\
		\bottomrule
	\end{tabular}
	}
\end{table}

\clearpage
\pagebreak

\section{Comparison with other gaussian based conditioning methods}
\begin{table}[!h]
	\caption{\textcolor{black}{Comparison of different methods to obtain the conditional means in the latent space of the VAE: (FPI) learns them using the fixed-point iteration proposed in the paper, whereas normal and uniform respectively sample them from a normal $\mathcal{N}(\bf{0}, \bf{I})$, or a uniform $\mathcal{U}(\bf{-1}, \bf{1})$ distribution. It can be noted that learning incremental means using FPI obtains  the highest performances on all benchmarks.}}
	\label{tab:ablation_gaussians_mean}
	\centering
	\resizebox{1.0\linewidth}{!}{%
	\begin{tabular}{lcccc} 
		\toprule
		Method to obtain the mean $\mu_{\y}$ & CUB200 & CARS196 & Img-R & CIFAR100 \\
		\midrule
		Multi-gaussian (learned using FPI) & $\mathbf{83.70_{\pm 0.23}}$ & $\mathbf{89.16_{\pm 0.23}}$ & $\mathbf{89.99_{\pm 0.08}}$ & $\mathbf{83.67_{\pm 0.25}}$ \\
		Multi-gaussian (sampled from a normal distribution $\mathcal{N}(\mathbf{0}, \mathbf{I})$) & $75.66_{\pm 0.47}$ & $83.42_{\pm 0.18}$ & $86.92_{\pm 0.94}$ & $77.44_{\pm 0.68}$ \\
		Multi-gaussian (sampled from a uniform distribution $\mathcal{U}(\mathbf{-1}, \mathbf{1})$) & $81.82_{\pm 0.47}$ & $87.72_{\pm 0.20}$ & $89.43_{\pm 0.21}$ & $83.02_{\pm 0.41}$ \\
		\bottomrule
	\end{tabular}
	}
\end{table}

\section{Backbones used in SOTA Comparison table}
\label{sup_sec:backbone_used_comparison_sota}
\begin{table}[!h]
	\centering
	\caption{Results from Table \ref{table:resutls_comparison} of the paper along with the ViT backbones used.}
	\label{table:sup_comparison_table_backbone_cgilcond}
	\resizebox{\linewidth}{!}{%
    \begin{tabular}{cccccc} 
		\toprule
		\textbf{Method}                & \textbf{CUB200}                     & \textbf{Cars196}                    & \textbf{Img-R}                      & \textbf{CIFAR100}                   & Backbone                   \\ 
		\hline
		L2P \cite{l2p}                 & 62.21\textsubscript{±1.92}          & 38.18\textsubscript{±2.33}          & 66.49\textsubscript{±0.40}          & 82.76\textsubscript{±1.17}          & \multirow{3}{*}{ViT-B/16}  \\
		DualPrompt \cite{dualprompt}   & 66.00\textsubscript{±0.57}          & 40.14\textsubscript{±2.36}          & 68.50\textsubscript{±0.52}          & 85.56\textsubscript{±0.33}          &                            \\
		CODA-Prompt \cite{codaprompt}  & 67.30\textsubscript{±3.19}          & 31.99\textsubscript{±3.39}          & 75.45\textsubscript{±0.56}          & 86.25\textsubscript{±0.74}          &                            \\ 
		\hline
		AttriCLIP \cite{attriclip}     & 50.07\textsubscript{±1.37}          & 70.98\textsubscript{±0.41}          & 86.25\textsubscript{±0.75}          & 81.40                               & ViT-L/14                   \\ 
		\hline
		MoE Adapters \cite{moe}        & 64.98\textsubscript{±0.29}          & 77.76\textsubscript{±1.02}          & \textbf{90.67\textsubscript{±0.12}} & 85.33\textsubscript{±0.21}          & \multirow{3}{*}{ViT-B/16}  \\ 
		\cmidrule{1-5}
		SLCA \cite{slca}               & 84.71\textsubscript{±0.40}          & 67.73\textsubscript{±0.85}          & 77.00\textsubscript{±0.33}          & \textbf{91.53\textsubscript{±0.28}} &                            \\
		SLCA++ \cite{slca++}           & \textbf{86.59\textsubscript{±0.29}} & 73.97\textsubscript{±0.22}          & 78.09\textsubscript{±0.22}          & 91.46\textsubscript{±0.18}          &                            \\ 
		\hline
		STAR-Prompt \cite{star-prompt} & 84.10\textsubscript{±0.28}          & 87.62\textsubscript{±0.20}          & 89.83\textsubscript{±0.04}          & 90.12\textsubscript{±0.32}          & ViT-L/14 +~ViT-B/16        \\ 
		\hline
		CGIL \cite{cgil}               & 83.12\textsubscript{±0.10}          & 89.27\textsubscript{±0.14}          & 89.42\textsubscript{±0.12}          & 86.02\textsubscript{±0.06}          & \multirow{4}{*}{ViT-L/14}  \\
		\textbf{VAE-FO (ours)}         & 83.70\textsubscript{±0.23}          & 89.16\textsubscript{±0.23}          & 89.99\textsubscript{±0.08}          & 83.67\textsubscript{±0.25}          &                            \\
		\textbf{VAE-FOD (ours)}        & 84.35\textsubscript{±0.45}          & 90.02\textsubscript{±0.19}          & 90.25\textsubscript{±0.16}          & 84.92\textsubscript{±0.14}          &                            \\
		\textbf{VAE-FODCE (ours)}      & 84.51\textsubscript{±0.24}          & \textbf{90.61\textsubscript{±0.07}} & 90.07\textsubscript{±0.10}          & 84.00\textsubscript{±0.37}          &                            \\
		\bottomrule
	\end{tabular}
	}
\end{table}

\section{Results without CLIP}
\label{seq:results_without_clip}
\begin{table}[!h]
	\centering
	\caption{Ablation as in Table~\ref{tab:ablation} of the paper, but using a ResNet18\cite{resnet} instead of a pretrained CLIP model, and the average incremental accuracy \cite{icarl} instead of the last accuracy. The ResNet18 is trained from scratch on CIFAR100, with either 50 classes or 40 classes in the first task, and then it encounters 5, 10, 20 or 60 subsequent tasks. \textcolor{black}{The feature extractor, which corresponds to the entire ResNet18 architecture without the classification layer, is trained in the first task only, and kept frozen in subsequent tasks.} The notation 50-T=5 stands for \emph{'50 classes in first task, and then 5 tasks'}. In the scenario 40-T=60, after the first task, the model encounters tasks with a single class. It can be noted that VAE-FO outperforms VAE-CEO on all benchmarks when using the orthogonality. Most interestingly, it reaches a better score in the more challenging and realistic setup 40-T=60 (1 class per task), than in 40-T=20.}
	\label{table:sup_ablation_again}
	\resizebox{\linewidth}{!}{%
	\begin{tabular}{cccccccc} 
		\toprule
		\multirow{2}{*}{Method}  & \multicolumn{3}{c}{Ablation Setting}                                                                                                                                                    & \multicolumn{4}{c}{CIFAR100}                                                                                           \\ 
		\cmidrule(l){2-8}
		& FPI conditioning        & \begin{tabular}[c]{@{}c@{}}Null space\\~optimization\\for the decoder\end{tabular} & \begin{tabular}[c]{@{}c@{}}Frozen decoder\\after first task\end{tabular} & 50-T=5                      & 50-T=10                     & 40-T=20                     & 40-T=60                      \\ 
		\midrule
		\multirow{2}{*}{VAE-CEO} & \multirow{2}{*}{\xmark} & \multirow{4}{*}{\xmark}                                                            & \xmark                                                                   & $58.88_\pm 1.03$            & $56.81_\pm 1.25$            & $51.63_\pm 1.96$            & $54.36_\pm 1.30$             \\
		&                         &                                                                                    & \cmark                                                                   & $61.81_\pm 0.91$            & $59.44_\pm 0.90$            & $53.36_\pm 1.16$            & $61.76_\pm 1.54$             \\ 
		\cmidrule(lr){1-2}\cmidrule(lr){4-8}
		\multirow{2}{*}{VAE-FO}  & \multirow{2}{*}{\cmark} &                                                                                    & \xmark                                                                   & $56.04_\pm 0.65$            & $44.74_\pm 0.25$            & $38.92_\pm 2.31$            & $41.61_\pm 0.50$             \\
		&                         &                                                                                    & \cmark                                                                   & $65.63_\pm 0.59$            & $64.67_\pm 0.61$            & $61.15_\pm 1.17$            & $65.63_\pm 1.27$             \\ 
		\midrule
		VAE-CEO                  & \xmark                  & \multirow{2}{*}{\cmark}                                                            & \multirow{2}{*}{\xmark}                                                  & $68.99_\pm 0.82$            & $66.85_\pm 0.74$            & $60.22_\pm 1.46$            & $60.14_\pm 1.48$             \\
		VAE-FO                   & \cmark                  &                                                                                    &                                                                          & $\mathbf{69.08_{\pm 0.71}}$ & $\mathbf{68.11_{\pm 0.66}}$ & $\mathbf{64.49_{\pm 1.35}}$ & $\mathbf{66.41_{\pm 1.44}}$  \\
		\bottomrule
	\end{tabular}
	}
\end{table}

\begin{table}[!h]
	\caption{Comparison of VAE-FO with FeTrIL \cite{fetril} implemented using PyCIL \cite{pycil} github with a ResNet18 \cite{resnet} backbone for both configurations. The ResNet18 is trained from scratch on CIFAR100, with either 50 classes or 40 classes in the first task, and then it encounters 5, 10, 20 or 60 subsequent tasks.  \textcolor{black}{The feature extractor, which corresponds to the entire ResNet18 architecture without the classification layer, is trained in the first task only and kept frozen in subsequent tasks.} The notation 50-T=5 stands for \emph{'50 classes in first task, and then 5 tasks'}. It can be noted that VAE-FO achieves the highest scores on all benchmarks, indicating the importance of learning a generative model instead of simply storing gaussian parameters as done in FeTrIL. The increased performance is most visible on scenarios with 1 class per task (40-T=60), where VAE-FO reaches up to 3.5 more accuracy points. VAE-FO + Resample means new synthetic data is sampled for each minibatch, while VAE-FO only samples 500 instances per class. The accuracy upper bound when training on all classes is $79.60_{\pm 0.19}$. The metric used is the average incremental accuracy  \cite{icarl}. }
	\label{table:sup_fetril_comparison}
	\centering
	\resizebox{0.8\linewidth}{!}{%
	\begin{tabular}{lcccc} 
		\toprule
		\multirow{2}{*}{Method} & \multicolumn{4}{c}{CIFAR100} \\
		\cmidrule(lr){2-5}
		& 50-T=5 & 50-T=10 & 40-T=20 & 40-T=60 \\
		\midrule
		FeTrIL & $68.40_{\pm 0.83}$ & $67.24_{\pm 0.84}$ & $64.30_{\pm 1.34}$ & $63.24_{\pm 1.33}$ \\
		\midrule
		VAE-FO (ours) & $69.08_{\pm 0.71}$ & $68.11_{\pm 0.66}$ & $64.49_{\pm 1.35}$ & $66.41_{\pm 1.44}$ \\
		VAE-FO + Resample (ours) & $\mathbf{69.34_{\pm 0.73}}$ & $\mathbf{68.41_{\pm 0.69}}$ & $\mathbf{64.77_{\pm 1.36}}$ & $\mathbf{66.74_{\pm 1.43}}$ \\
		\bottomrule
	\end{tabular}
	}
\end{table}

\section{Computational details} \label{sec:computational_details_cgilcond}
\textcolor{black}{All the experiments in this paper were carried on single V100 GPU with 16GB or 32GB or RAM. The compute time for each dataset and each experiment is as following.
	\begin{itemize}
		\item \emph{CIFAR100}: 6 hours for a single run totaling 18 hours for 3 seed experiments.
		\item \emph{CUB200}: 17 hours for a single run totaling 51 hours for 3 seed experiments.
		\item \emph{Cars196}: 17 hours for a single run totaling 51 hours for 3 seed experiments.
		\item \emph{ImageNet-R}: 17 hours for a single run totaling 51 hours for 3 seed experiments.
	\end{itemize}
	Reproducing the experiments in the paper would require approximately 2500 GPU hours. \\
	The total estimated GPU hours used for this paper is approximately 10000 (including experiments that did not make it in the paper / parameter optimization). }

\section{Limitations} \label{sec:limitations_cgilcond}
\textcolor{black}{As the number of tasks tends toward infinity, the null space optimization will inevitably saturate which will cause forgetting in the incremental decoder network. One solution to fix this issue, would consist in initializing a new decoder network once the null space saturates, according to a predefined threshold on the \emph{"proportion"} ratio defined in section~\ref{seq:orthogonality implemetation details}. Nonetheless it should be noted that we did not observe significant forgetting, even on challenging scenarios with 60 total task available in Table~\ref{table:sup_fetril_comparison} of the Appendix, so the method still works with many tasks.}

\raggedbottom

\pagebreak
\clearpage

\section{Fixed-Point Iteration Proof}
\label{seq:fpi_proof}
In this section, we show the derivation steps to obtain Eq.~\ref{eq:fpi} from the paper. We start by deriving the equations of the optimality conditions for the likelihood term in Eq.~\ref{sec:optim_likelihood} and then for the KLD term in Eq.~\ref{eq:kld_sum_all}. Then we sum and simplify them, to obtain the final result in Eq.~\ref{eq:fpi_final_sup_mat}. 
\subsection{Derivation of the Likelihood term}

The optimality conditions for the negative log likelihood of $\mu_{\y}$ is shown in Eq.~\ref{sec:optim_likelihood}.
\begin{equation}
	\label{sec:optim_likelihood}
	\frac{\partial}{\partial \mu_{\y}} (\sum_{\substack{\x,\y \in \mathcal{D}^{t}}} -\log (\mathcal{N}(\mu_{\phi, \x, \y}; {\mu_{\y}, \bf{I}}))) = -\frac{1}{N_{\y}}\sum_{i=1}^{N_{\y}}  (\mu_{\phi, \x_{i}, \y_{i}} - \mu_{\y}) = 0
\end{equation}

\noindent By isolating $\mu_{\y}$ we would obtain the result in Eq.~\ref{sec:optim_likelihood_res}, which corresponds to the mean of the data in the latent space, where $N_{\y}$ is the number of images with label $\y$. However this formulation does not take into account the KLD term that we define in the next section.
\begin{equation}
	\label{sec:optim_likelihood_res}
	\mu_{\y} = \frac{1}{N_{\y}}  \sum_{i=1}^{N_{y}} \mu_{\phi, \x_{i}, \y_{i}}
\end{equation}

\subsection{Derivation of the KLD term}
\noindent We remind that the KLD between two gaussians $\mathcal{N}_{\y}$ and $ \mathcal{N}_{\y'}$ in Eq.~\ref{eq:kld_formula}
\begin{equation}
	\label{eq:kld_formula}
	KLD(\mathcal{N}_{\y} || \mathcal{N}_{\y'}) = \frac{1}{2}\big[ 
	\log \frac{|\Sigma_{\y'}|}{|\Sigma_{\y}|} - d + \Tr({\Sigma_{\y'}^{-1} \Sigma_{\y}}) + (\mu_{\y'} - \mu_{\y})^{\top} \Sigma_{\y'}^{-1} (\mu_{\y'} - \mu_{\y})
	\big]
\end{equation}

\noindent  By setting $\Sigma_{\y'} = \Sigma_{\y} = \mathbf{I}$;\;\; Eq.~\ref{eq:kld_formula} simplifies to
\begin{equation}
	KLD(\mathcal{N}_{\y} || \mathcal{N}_{\y'}) = \frac{1}{2} 
	(\mu_{\y'} - \mu_{\y})^{\top} (\mu_{\y'} - \mu_{\y})
\end{equation}

\noindent  \textcolor{black}{To find the expression of the optimality conditions for a mean $\mu_{\y}$, we express it as function of the partial derivatives of all pairwise KLD where the term $\mu_{\y}$ is taken into account in the loss. This includes all the pairs $KLD(\mathcal{N}_{\y} || \mathcal{N}_{\y'}) $ and their reverse $KLD(\mathcal{N}_{\y'} || \mathcal{N}_{\y})$ that we sum together. However since in this special case the KLD is symmetric, we may directly write:}
\begin{equation}
	\frac{\partial}{\partial \mu_\y} (-\log (KLD(\mathcal{N}_{\y} || \mathcal{N}_{\y'}))) = \frac{\partial}{\partial \mu_\y} (-\log (KLD(\mathcal{N}_{\y'} || \mathcal{N}_{\y})))  = 2 \frac{(\mu_{\y'} - \mu_{\y})}{\lVert \mu_{\y'} - \mu_{\y} \rVert^{2}_{2}}
\end{equation}	

\noindent Since the KLD term is weighted with a scalar $\lambda$ in Eq.~\ref{eq:fpi_full} of the paper, we only need a proportionality relation as in Eq.~\ref{eq:kld_proporitonal_lambda}.

\begin{equation}
	\label{eq:kld_proporitonal_lambda}
	\frac{\partial}{\partial \mu_\y} (-\log (KLD(\mathcal{N}_{\y} || \mathcal{N}_{\y'}))) + \frac{\partial}{\partial \mu_\y} (-\log (KLD(\mathcal{N}_{\y'} || \mathcal{N}_{\y})))  \propto  \lambda \frac{(\mu_{\y'} - \mu_{\y})}{\lVert \mu_{\y'} - \mu_{\y} \rVert^{2}_{2}}
\end{equation}	

\noindent By summing on all other gaussians (excluding $\mathcal{N}_{\y}$) starting from Eq.~\ref{eq:kld_proporitonal_lambda}, we obtain the results in Eq.~\ref{eq:kld_sum_all} for the optimality conditions of $\mu_{\y}$ with respect to the KLD term.

\begin{equation}
	\label{eq:kld_sum_all}
	\lambda \sum_{\substack{\y' \in \Y^{1:t} \\ \y'\neq \y}} \frac{(\mu_{\y'} - \mu_{\y})}{\lVert \mu_{\y'} - \mu_{\y} \rVert^{2}_{2}} = 0
\end{equation}

\subsection{Summation of both terms}
By summing both terms from Eq.~\ref{sec:optim_likelihood}~and~\ref{eq:kld_sum_all} we obtain  Eq.~\ref{eq:fpi_sum}.
\begin{equation}
	\label{eq:fpi_sum}
	\underbrace{-\sum_{i=1}^{N_{\y}}  (\mu_{\phi, \x_{i}, \y_{i}} - \mu_{\y}) }_{\text{likelihood term}}  
	+ \underbrace{\lambda \sum_{\substack{\y' \in \Y^{1:t} \\ \y'\neq \y}} \frac{(\mu_{\y'} - \mu_{\y})}{\lVert \mu_{\y'} - \mu_{\y} \rVert^{2}_{2}}}_{\text{KLD term}} = 0 
\end{equation}

\begin{equation}
	\label{eq:fpi_sum_mult}
	\underbrace{\sum_{i=1}^{N_{\y}}  (\mu_{\phi, \x_{i}, \y_{i}} - \mu_{\y}) }_{\text{likelihood term}}  
	+ \underbrace{\lambda  \sum_{\substack{\y' \in \Y^{1:t} \\ \y'\neq \y}} \frac{(\mu_{\y} - \mu_{\y'})}{\lVert \mu_{\y} - \mu_{\y'} \rVert^{2}_{2}}}_{\text{KLD term}} = 0 
\end{equation}

\noindent In Eq.~\ref{eq:fpi_sum_mult}, since we have two terms with different summations (likelihood term and KLD term), we have to normalize them individually to balance them. \\
\noindent For the likelihood term this involves dividing by $N_{\y}$ (the number of images with label $\y$)  to get the mean.\\
\noindent For the KLD term, this involves diving by the number of KLD pairs $|\Y^{1:t}| \cdot (|\Y^{1:t}| - 1)$. 

\begin{equation}
	\label{eq:fpi_normalize_before}
	\frac{1}{N_{\y}}\sum_{i=1}^{N_{\y}}  (\mu_{\phi, \x_{i}, \y_{i}}) - \mu_{\y} 
	+ \frac{\lambda}{|\Y^{1:t}| \cdot (|\Y^{1:t}| - 1)} \sum_{\substack{\y' \in \Y^{1:t} \\ \y'\neq \y}} \frac{(\mu_{\y} - \mu_{\y'})}{\lVert \mu_{\y} - \mu_{\y'} \rVert^{2}_{2}} = 0 
\end{equation}

However to avoid cluttered notations, we absorb the term  $|\Y^{1:t}| \cdot (|\Y^{1:t}| - 1)$ in $\lambda$ as in Eq.~\ref{eq:fpi_normalize}, and obtain the result from Eq.~\ref{eq:fpi_without_tau}.

\begin{equation}
	\label{eq:fpi_normalize}
	\frac{1}{N_{\y}}\sum_{i=1}^{N_{\y}}  (\mu_{\phi, \x_{i}, \y_{i}}) - \mu_{\y} 
	+ \lambda \sum_{\substack{\y' \in \Y^{1:t} \\ \y'\neq \y}} \frac{(\mu_{\y} - \mu_{\y'})}{\lVert \mu_{\y} - \mu_{\y'} \rVert^{2}_{2}} = 0 \quad\text{(absorb normalization in $\lambda$)}
\end{equation}

\begin{equation}
	\label{eq:fpi_without_tau}
	\mu_{\y} = \frac{1}{N_{\y}}\sum_{i=1}^{N_{\y}}  (\mu_{\phi, \x_{i}, \y_{i}})  
	+ \lambda  \sum_{\substack{\y' \in \Y^{1:t} \\ \y'\neq \y}} \frac{(\mu_{\y} - \mu_{\y'})}{\lVert \mu_{\y} - \mu_{\y'} \rVert^{2}_{2}}  \quad\text{(isolate a term $\mu_{\y}$)}
\end{equation}

\noindent We obtain the final result from Eq.~\ref{eq:fpi_final_sup_mat} by adding the time steps $\tau$ in Eq.~\ref{eq:fpi_without_tau}, indeed we express $\mu_{\y}$ at iteration ($\tau$), as a function of other means (including $\mu_{\y}$) at iteration ($\tau$ - 1) 

\begin{equation}
	\label{eq:fpi_final_sup_mat}
	\mu_{\y}^{(\tau)} = \frac{1}{N_{\y}}\sum_{i=1}^{N_{\y}}  (\mu_{\phi, \x_{i}, \y_{i}})  
	+ \lambda  \sum_{\substack{\y' \in \Y^{1:t} \\ \y'\neq \y}} \frac{(\mu_{\y}^{(\tau - 1)} - \mu_{\y'}^{(\tau - 1)})}{\big \lVert \mu_{\y}^{(\tau - 1)}  - \mu_{\y'}^{(\tau - 1)}  \big \rVert^{2}_{2}}  \quad\text{(include time steps $\tau$)}
\end{equation}

\end{document}